\documentclass{article}

\usepackage[preprint]{neurips_2025}

\usepackage[utf8]{inputenc} 
\usepackage[T1]{fontenc}    
\usepackage{hyperref}       
\usepackage{url}            
\usepackage{booktabs}       
\usepackage{amsfonts}       
\usepackage{nicefrac}       
\usepackage{microtype}      

\usepackage{pifont}
\usepackage{graphicx}
\usepackage{subcaption}
\usepackage{inconsolata}
\usepackage[most]{tcolorbox}
\usepackage{enumerate}

\usepackage{xcolor} 
\usepackage{pgffor} 
\usepackage{placeins} 
\usepackage{textcomp}
\usepackage{utfsym}
\usepackage{makecell}
\usepackage{amsmath}
\usepackage{wrapfig}

\makeatletter
\newcommand{\rmnum}[1]{\romannumeral #1}
\newcommand{\Rmnum}[1]{\expandafter\@slowromancap\romannumeral #1@}
\makeatother

\newcommand{\eg}{{\emph{e.g.}}, }

\definecolor{calloutcolor}{HTML}{D6EDFF} 
\tcbset{ callout/.style={ enhanced, colback=calloutcolor!40, 
colframe=calloutcolor!40, 
boxrule=0mm, 
colbacktitle=calloutcolor, 
coltitle=black, 
arc=2mm, 
width=\linewidth, 
boxsep=2mm, 
left=1mm, right=1mm, top=1mm, bottom=1mm, 
fonttitle=\bfseries, 
title={#1}, 
toptitle=0mm, 
bottomtitle=0mm, 
} }

\title{Predictable Scale: Part I, Step Law –  
Optimal Hyperparameter Scaling Law in Large Language Model Pre-training}

\author{
    Houyi Li\thanks{Equal contribution.} \\
    StepFun, Fudan University \\
\And
    Wenzhen Zheng$^{\ast}$ \\
    StepFun \\
\And
    Qiufeng Wang$^{\ast}$ \\
    StepFun \\
\AND
    Hanshan Zhang \\
    StepFun \\
\And
    Zili Wang \\
    StepFun \\
\And
    Shijie Xuyang \\
    StepFun, Fudan University \\
\AND
    YuanTao Fan \\
    StepFun \\
\And
    Zhenyu Ding \\
    Xi'an Jiaotong University \\
\And
    Haoying Wang \\
    Xi'an Jiaotong University \\
\AND
    Ning Ding \\
    Xi'an Jiaotong University \\
\And
    Shuigeng Zhou \\
    Fudan University \\
\And
    Xiangyu Zhang \\
    StepFun, Megvii Technology \\
\And
    Daxin Jiang \\
    StepFun \\
}

\begin{document}

\maketitle

\begin{abstract}

    The impressive capabilities of Large Language Models (LLMs) across diverse tasks are now well\text{-}established, yet their effective deployment necessitates careful hyperparameter optimization. 
    Although existing methods have explored the influence of hyperparameters on model performance, a principled and generalizable framework across model architectures and data recipes remains absent.
    In this study, we conduct an unprecedented empirical investigation\text{-} training over 3,700 LLMs from scratch across 100 trillion tokens, consuming nearly one million NVIDIA H800 GPU hours to establish a universal Scaling Law for hyperparameter optimization in LLM Pre-training, called \textbf{Step Law}. 
    We empirically observe that, under fixed model size ($N$) and dataset size ($D$), the hyperparameter landscape exhibits convexity with a broad optimum, substantially reducing the complexity of hyperparameter search. 
    Building on this insight, we formally define and empirically validate the Step Law:
    The optimal learning rate follows a power-law relationship with $N$ and $D$, while the optimal batch size is primarily influenced by $D$ and remains largely invariant to $N$.
    Notably, our estimated optima deviate from the global best performance found via exhaustive search by merely \textbf{0.094\%} on the test set.
    To our best known, Step Law is the \textbf{first} that unifies different model shapes and structures, such as Mixture-of-Experts models and dense transformers, as well as establishes optimal hyperparameter scaling laws across diverse data recipes. 
    We contribute a universal, plug-and-play optimal hyperparameter tool for the community, which is expected to advance efficient LLM training at scale. 
    All experimental code, data and checkpoints are publicly available at \href{https://github.com/step-law/steplaw}{https://github.com/step-law/steplaw}. 
\end{abstract}

\newpage

\section{Introduction}
\label{section:intro}

State-of-the-art Large Language Models (LLMs) have surged to unprecedented scales, training on billions of parameters and trillions of tokens \cite{Brown2020,Jin2023,biderman2023pythiasuiteanalyzinglarge,workshop2023bloom176bparameteropenaccessmultilingual,Touvron2023a,Touvron2023b,Grattafiori2024,DeepSeek-AI2024,Yang2024,DeepSeek-AI2024b,DeepSeek-AI2025}. For example, Llama 3 \cite{Grattafiori2024} was trained on 15 trillion tokens, underscoring the escalating computational demands. 
At this scale, training hyperparameters, particularly the learning rate (LR) and batch size (BS), is extremely critical. Excessively large learning rates can cause training to diverge, whereas insufficient learning rates will result in slow convergence \cite{Shen2024,Wen2024}.
Likewise, batch size must strike a balance between throughput and generalization \cite{Perko2023,Filatov2024,McCandlish2018} in LLM pre-training. 
Traditional grid search is prohibitively expensive, prompting transfer methods that extrapolate optimal settings from smaller models to large ones \cite{Yang2020,Yang2023}.

\begin{table*}[t!] 
    \renewcommand{\arraystretch}{1.3}

    \resizebox{\textwidth}{!}{%
    \centering
    \Large
    
     \begin{tabular}{lccccccccc}
        \toprule[1.8pt] \textbf{Name} & \textbf{Data Recipe}       & \textbf{Model Sparsity}      & \textbf{Learning Rate}                              & \textbf{Batch Size}                   & \textbf{Relative Error}       \\
        \hline
        
        OpenAI Law \cite{kaplan_scaling_2020}& \color{red}\usym{2718}   & \color{red}\usym{2718}   & $3.239*10^{-3}+-1.395*10^{-4}log(N)$              & $2e18{\mathcal{L}^{-4.76190}}$          & 9.51\textperthousand          \\
        
        Microsoft Law  \cite{Bjorck2024}  &  \color{red}\usym{2718}     & \color{red}\usym{2718}  & $1.3192e^{-5}N^{-0.23}D^{-0.32}$       & -           & 9.25\textperthousand          \\
        DeepSeek Law     \cite{DeepSeek-AI2024}&  \color{red}\usym{2718}   & \color{red}\usym{2718}     & $0.3188C^{-0.1250}$                      & $0.2920C^{0.3271}$            & 9.26\textperthousand          \\
        Porian Law    \cite{Porian2024}   & \color{red}\usym{2718}    & \color{red}\usym{2718}       & $3.7N^{-0.36}$                           & $0.7576N^{0.703}$             & 3.71\textperthousand          \\
        MiniCPM Law     \cite{Hu2024}     & \color{red}\usym{2718}      & \color{red}\usym{2718}       & -                                 & $\frac{2e18}{L^{6.24}}$       & -                             \\
        \makecell[l]{MeiTuan Law      \cite{Wang2024}}   & \color{red}\usym{2718}      & \color{green}\usym{2714}     & $\lambda{\mathcal{L}^{-\alpha}}$                   & $\lambda_{B}{\mathcal{L}^{-\alpha_{B}^{-1}}}$ & -                             \\
        \hline
         Ours (Step Law)             &     \color{green}\usym{2714}  & \color{green}\usym{2714}   & $1.79N^{-0.713}D^{0.307}$                & $0.58D^{0.571}$               & \textbf{0.94\textperthousand} \\
        \bottomrule[1.2pt]
    \end{tabular}%
    }
    
    \caption{Comparison of optimal hyperparameter scaling laws across
    different approaches. \textbf{Data Recipe} and \textbf{Model Sparsity} denotes whether the approach is suitable for different data recipe and model sparsity. \textbf{Relative Error} denotes the relative loss, as same as Fig.~\ref{fig:firt_pic_baselines_compare}. The variables in scaling laws are described  in Sec.~\ref{sec:notation}.}
    \label{tab:related_work}
    \vskip -20pt
\end{table*}

To avoid exhaustive searches, the community has advanced hyperparameter-transfer rules. $\mu$P \cite{Yang2022} first established learning rate scaling with model width, later extended to depth and other architectures \cite{Everett2024,Lingle2024,Blake2024}.
Empirical scaling laws then linked learning rate and batch size to model size \cite{kaplan_scaling_2020}, refined for dense models in subsequent work \cite{Bjorck2024,DeepSeek-AI2024,Porian2024,Hu2024}. Recently, studies have explored these interactions in Mixture-of-Experts (MoE) models \cite{Du2021,Fedus2021,Wang2024,Ludziejewski2025}.

\begin{wrapfigure}{r}{7.1cm}
    \begin{center}
        \centerline{\includegraphics[trim=0 0 0 32, clip, width=0.53\textwidth]{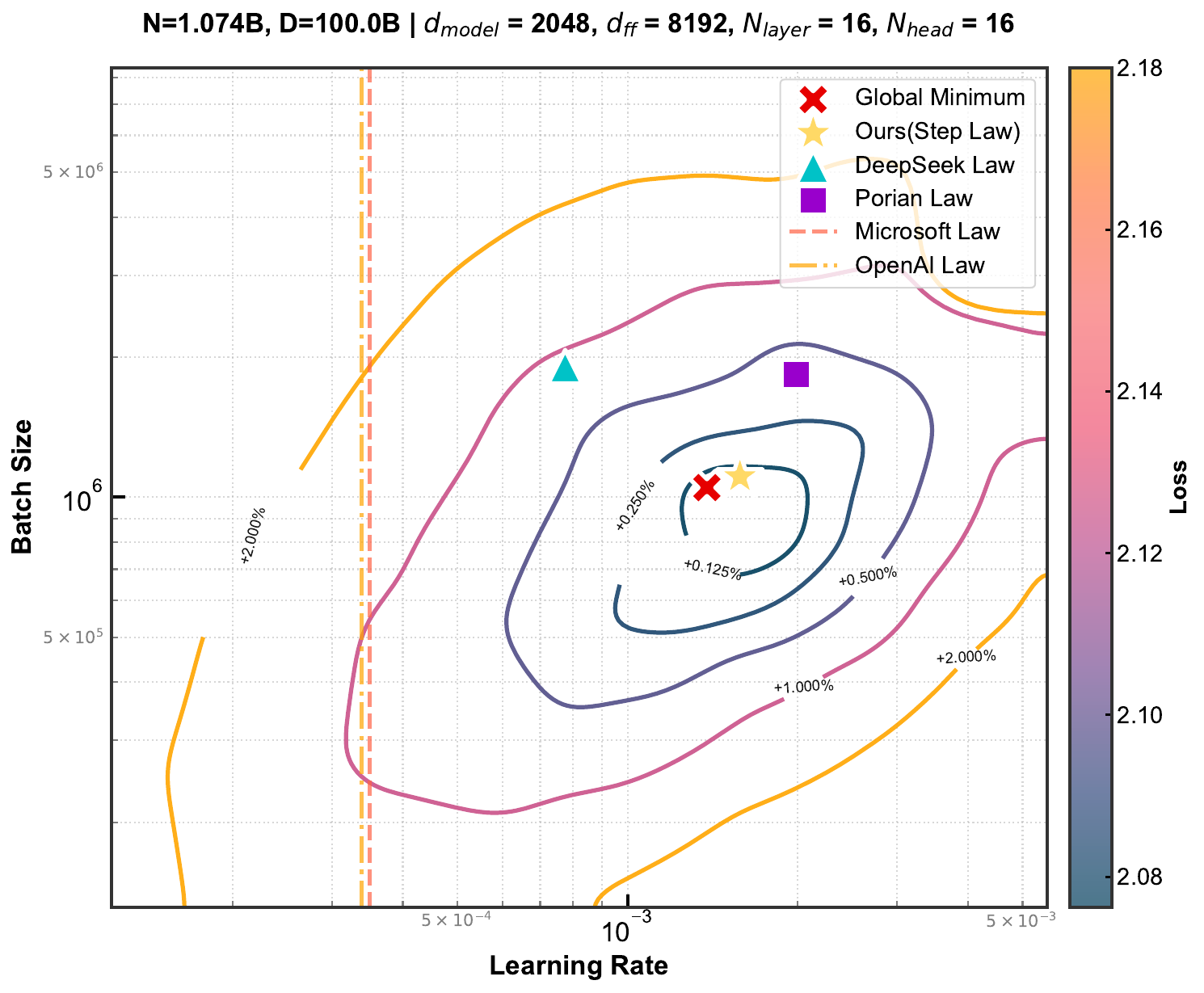}}
        \vskip 10pt
        \caption{This plot shows the hyperparameter space for a 1B model trained on 100B tokens. We trained 120 LLMs from scratch with different LR and BS combinations, obtaining contour lines and global optimal points based on real data. Optimal points represent the lowest training loss for each LR and BS pair, while contour lines depict the relative loss differences from these optima. Step Law predicts the optimum with the highest accuracy compared to other methods, nearly matching the global optimal points.}
        \label{fig:firt_pic_baselines_compare}
        \vskip -28pt
    \end{center}
\end{wrapfigure}

However, a significant gap remains in understanding hyperparameter transfer across different dimensions: data recipe, model shape, model sparsity, and dataset sizes $D$.
Although existing research has tried to understand scaling behavior across model sizes \cite{kaplan_scaling_2020,Halfon2024}, the interaction of these other critical factors remains under-explored.
As illustrated in Tab.~\ref{tab:related_work}, existing scaling laws almost uniformly fail to accommodate variations in data recipe and model sparsity.
In contrast, our work is the only approach that simultaneously supports diverse data recipes and sparsity levels and reduces the relative error to merely 0.94\textperthousand, with a significant improvement over the prior methods.
To address this gap, we derive universal hyperparameter scaling laws that span all key dimensions of LLM pre-training and guide optimal parameter selection.
Our results show that a single empirically obtained law generalizes across model shape, sparsity, data recipe and dataset size, improving both accuracy and applicability.

Our main contributions are as follows:

(\rmnum{1}) This paper establish the first universal and robust Scaling Law for hyperparameter optimization in LLM Pre-training, called \textbf{Step Law}. 
We discover the power-law relationship between optimal learning rate $\eta(N, D)$ and optimal batch size $B(D)$. 
Step Law demonstrates that the optimal batch size exhibits a primary dependence on dataset size $D$, while the optimal learning rate manifests a joint dependence on both model parameters $N$ and dataset size $D$:

\vskip -10pt
\begin{equation}
\begin{aligned}
    \eta(N, D) & = 1.79N^{-0.713}D^{0.307}  \\
    B(D)       & = 0.58D^{0.571}           
    \label{stepai}
\end{aligned}
\end{equation}
\vskip -8pt

Step Law achieves superior convergence results compared to baseline methods when generalized to 1B models, as shown in Fig.~\ref{fig:firt_pic_baselines_compare}.
The hyperparameters predicted by Step Law yield a test-set loss within just 0.094\% above the global optimum found via exhaustive search, significantly outperforming all other methods.
It provides a plug-and-play formula that significantly minimizes hyperparameter tuning efforts, making it highly practical for industrial-scale training.

\begin{figure}[!t]
    \setlength{\abovecaptionskip}{0pt}
    \setlength{\belowcaptionskip}{10pt}
    \begin{center}
        \centerline{\includegraphics[width=1.0\textwidth, trim=0 -50 0 0, clip]{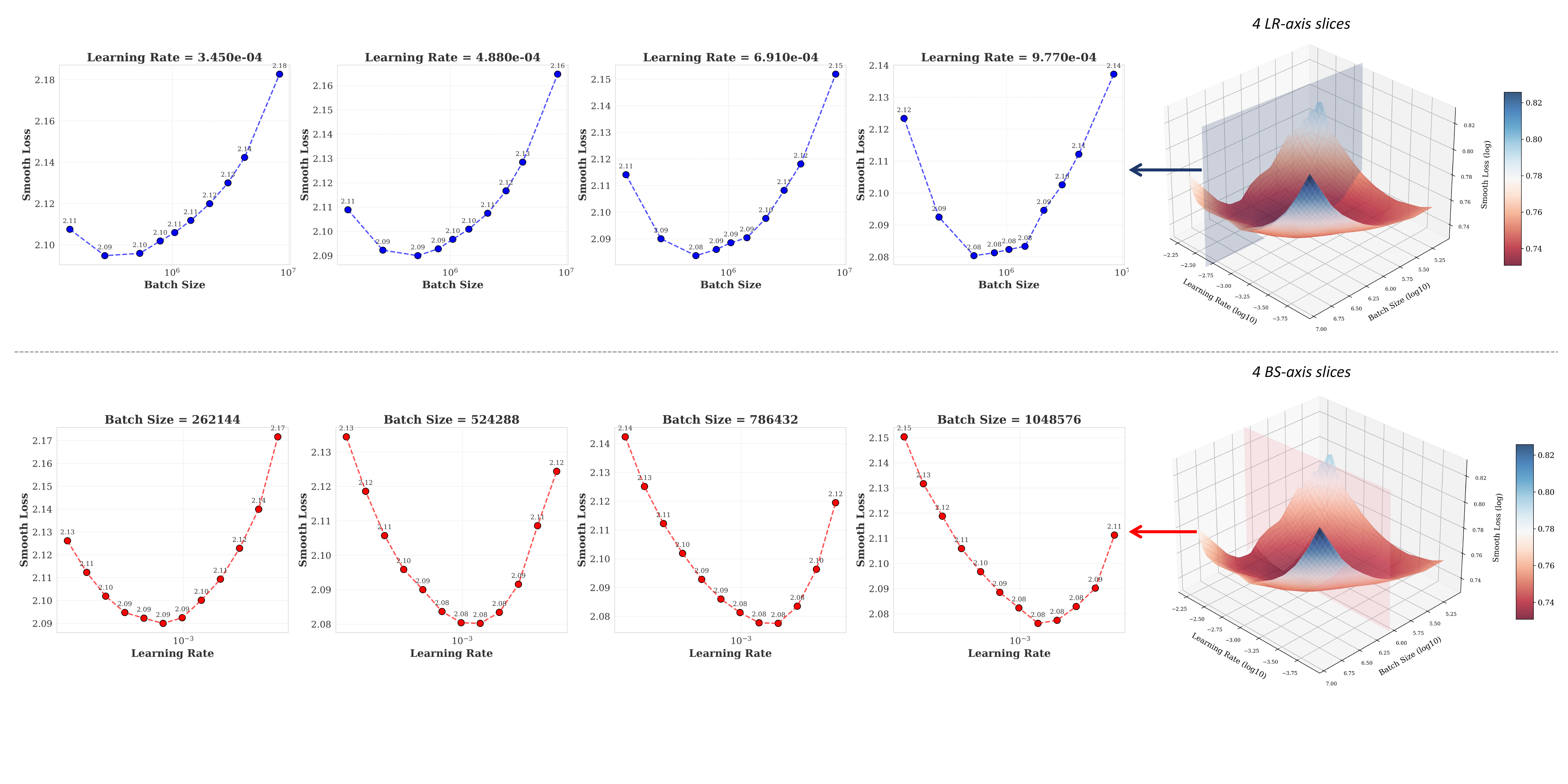}}
        \caption{Learning Rate vs. Batch Size Loss Landscape Analysis for 1B Model (Trained on 100B Tokens): Scatter Plots and 3D Surface Visualizations of Hyperparameter Sensitivity.}
        \label{fig:Convex}
    \end{center}
    \vskip -40pt
\end{figure}

(\rmnum{2}) To the best of our knowledge, we are the first to discover and demonstrate the convexity property of the loss landscape under fixed parameter count and dataset size conditions. This significant preliminary provides fundamental insights into hyperparameter optimization, as shown in Fig.~\ref{fig:Convex}.

(\rmnum{3}) 
Step Law is the first to investigate the transferability and robustness of optimal hyperparameter scaling laws across diverse pre-training data recipes and model architectures. 
We find that Step Law exhibits strong invariance across dense and sparse (MoE) LLMs with varying sparsity levels, and generalizes well across data recipes, model architectures, and sparsity configurations, as confirmed by extensive grid search.
In addition, we evaluate Step Law on a 6.5B MoE model trained across varying dataset sizes. In all  settings, the learning rate and batch size predicted by Step Law yield the lowest validation smooth loss, outperforming other methods.

(\rmnum{4}) We conduct an unprecedented large-scale empirical study, comprising 3,700 full LLM training runs across Dense and MoE models (with varying sparsity levels), diverse architectures, data recipes, and hyperparameter settings. The total compute consumed approaches 1 million H800 GPU hours (equivalent to over one million dollars), processing approximately 100 trillion tokens. To the best of our knowledge, this represents the largest dataset of hyperparameter optimization results in the field, derived entirely from empirical observations without relying on prior assumptions. All training checkpoints and hyperparameter configurations are publicly available.

\section{Related Works}

Hyperparameter transfer extrapolates optimal settings from smaller to larger models and is essential for efficient large-scale LLM training.
In particular, learning rate and batch size have a major impact on pre-training performance \cite{Halfon2024}.
Existing work divides into theory-driven and data-driven methods.

Theory-driven methods began with $\mu \text{P}$\text{'}s learning rate rules for varying model widths \cite{Yang2022}. Subsequent work extended these rules to model depth and other architectural variants \cite{Everett2024,Lingle2024,Blake2024,Bordelon2023}. 
All of these approaches require custom initialization or modified attention mechanisms. \textbf{They do not address variations in data recipe, sparsity level or dataset size, nor do they predict batch size.}

Data-driven methods express hyperparameters as functions of model size $N$ and dataset size $D$. \cite{kaplan_scaling_2020} first related learning rate to $N$ , and later work incorporated $D$ into a joint law $LR(N,D)=CN^{-\alpha}D^{-\beta}$ \cite{Bjorck2024}.
Batch size rules have been proposed based on expected loss \cite{Wang2024,Hu2024} or two-dataset fits \cite{Porian2024}, but these either require prior loss estimates, ignore $N$’s influence on learning rate, or assume fixed compute budgets \cite{DeepSeek-AI2024}. 
Early studies also show that optimal batch size depends mainly on $D$ rather than $N$ \cite{McCandlish2018,Huawei-Noah-Ark-Lab2024,zhang2024}.

In contrast, our Step Law demonstrates that $N$ and $D$ alone suffice to predict both learning rate and batch size. We validate these scaling laws across model shapes, sparsity levels and data recipes, offering unified and accurate hyperparameter guidance for LLM pre-training.

\section{Methodology}

\subsection{Problem Setup}
\label{preliminaries}
For training LLMs, the comprehensive performance
metric is defined as
\begin{equation}
    \mathcal{L}(\mathbb{A}, \mathbb{D}, N, D, \text{LR}, \text{BS}),
\end{equation}
where $\mathbb{A}$, $\mathbb{D}$, $N$, $D$, \text{LR}, and \text{BS} represent
the model architecture space, training data distribution, number of non-vocabulary
    parameters, number of training tokens, learning rate and batch size.

Based on this definition, when considering specific conditions, first, given
that both $\mathbb{A}$ and $\mathbb{D}$ are discrete variables, the performance
metric can alternatively be expressed as

\begin{equation}
    \mathcal{L}_{\mathbb{A},\mathbb{D}}(N, D, \text{LR}, \text{BS}). \label{eq:fix_A_D}
\end{equation}

Furthermore, for given $N$ and $D$, Eq.\eqref{eq:fix_A_D} can be transformed
into
\begin{equation}
    \begin{aligned}
            & \mathcal{L}_{\mathbb{A}, \mathbb{D}, N, D}(\text{LR}, \text{BS})
    \end{aligned}
\end{equation}
In light of the above transformations, we can generate the following
definition.

\textbf{Definition 1: }\label{defin:1} (\emph{Hyperparameter Optimality}) For fixed architecture $\mathbb{A}$, data distribution $\mathbb{D}$, and training
budget $(N, D)$, the optimal learning rate $\eta$ and batch size $B$ satisfy:
\begin{equation}
    \eta, B = \mathop{\mathrm{arg\,min}}_{\text{LR}, \text{BS}}\mathcal{L}_{\mathbb{A},\mathbb{D},N,D}
    (\text{LR}, \text{BS}).
\end{equation}

\subsection{Experimental Settings}
\label{exp_set}
We train our models using a language modeling loss on a dataset comprising web text, math, and code, with proportions aligned with those in Llama ~\cite{Touvron2023a,Touvron2023b} (see Tab.~\ref{tab:data_recipes} for details). The
dataset is tokenized using a BPE \cite{Gage1994ANA} tokenizer with a vocabulary size of 65,536.

Following the prevailing configuration in recent large-scale models, our model architecture uses RMSNorm \cite{Zhang2019} for pre-normalization and the SwiGLU \cite{Shazeer2020} activation function in the feed-forward network, without applying dropout \cite{JMLR:v15:srivastava14a}.
We mainly use ALiBi \cite{Press2021} positional encoding.  
The models are initialized from scratch,
with weights sampled from a truncated normal distribution (mean of 0, standard
deviation of 0.02). 

We use the AdamW \cite{Loshchilov2017} optimizer with $\beta$ values of [0.9, 0.95], an epsilon of
$10^{-8}$, a weight decay of 0.1, and a gradient clipping norm of 1.0. Our learning rate schedule includes a linear warmup phase over the initial 2,000 steps, followed by a cosine decay reaching a final learning rate of $10^{-5}$ for the remainder of the training. A detailed analysis and rationale for this strategy are provided in Sec.~\ref{fixed_final_lr}. The sequence length is set to 2,048 tokens. The learning rate is selected from a logarithmic sequence of powers of 2, spanning exponents from -10.5 to -7.0 in increments of 0.5. 
The batch size is selected from a geometric progression, ranging from 32{,}768 to 4{,}194{,}304, with each subsequent value being $\sqrt{2}$ times the previous one.
These parameter configurations correspond to the 18 LLMs detailed in Tab.~\ref{tab:struct} in Appendix~\ref{app:struct}.

\subsection{Preliminary Experiments}

\label{Hypothesis_Validation}
\subsubsection{Loss Landscape Convexity Analysis}


Through extensive empirical analysis, we identify a fundamental property
of the loss landscape with respect to hyperparameters: both the learning rate and
batch size exhibit convex relationships with the training loss under fixed model
parameters and dataset size conditions. Fig.~\ref{fig:Convex} illustrates a representative experimental result, with comprehensive results further elaborated in Appendix~\ref{app:convex}.


Furthermore, we observe that the loss surface demonstrates a stable region around
the optimal configuration, evidenced by the plateau-like behavior shown in
Fig.~\ref{fig:loss_bpc_heatmaps}. This stability provides practical
tolerance for small deviations in hyperparameter selection while maintaining
near-optimal performance.
These properties form the theoretical foundation for our subsequent
development of scaling laws and validate their applicability across
different architectural configurations.

\subsubsection{Impact of Final Learning Rate Schedule}
\label{fixed_final_lr}


\begin{wrapfigure}{r}{7.1cm}
    \centering
    \vskip -8pt
    \includegraphics[width=0.5\textwidth]{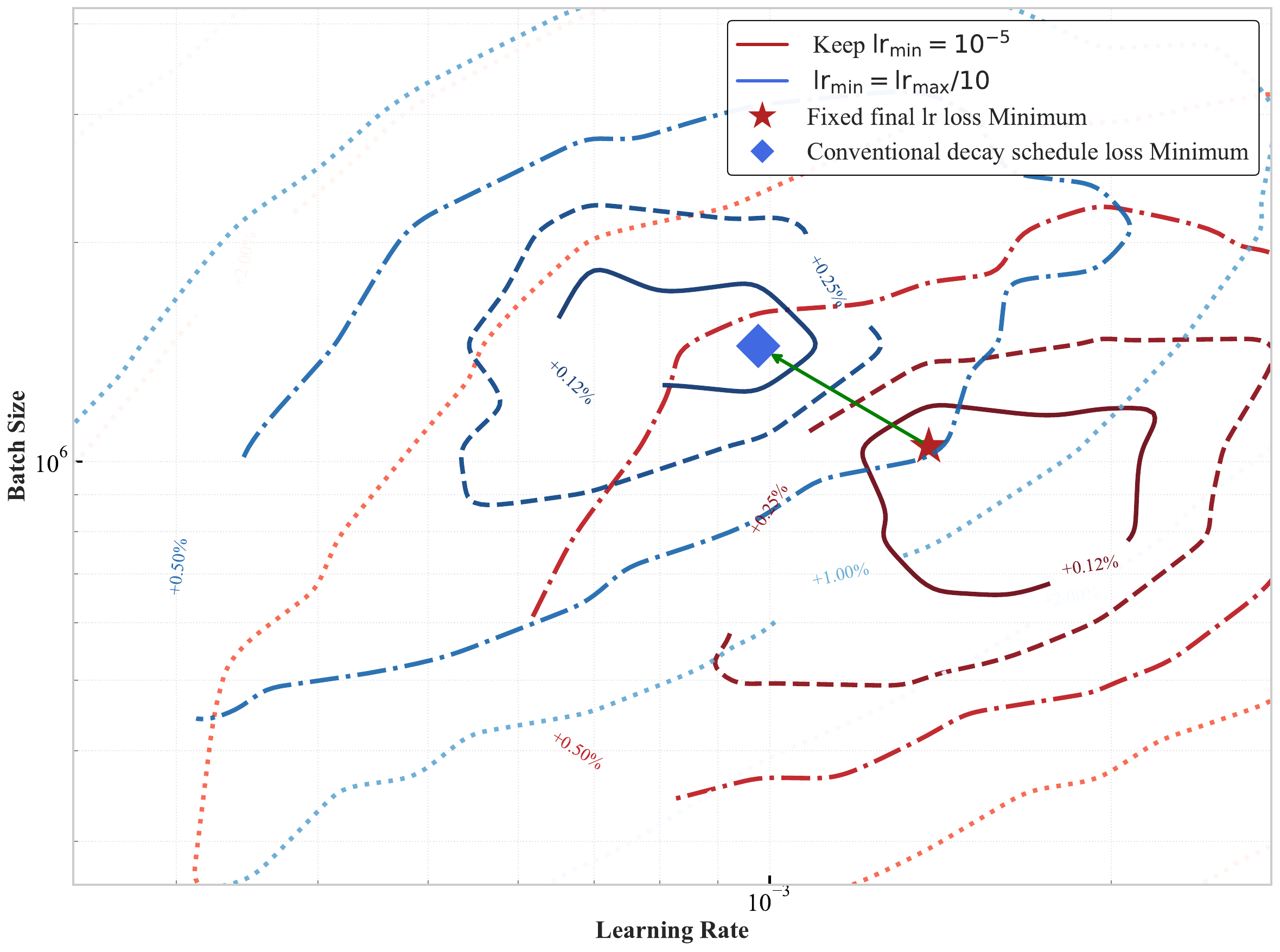}
    \caption{\textbf{Comparison of learning rate schedules.} Contour plots illustrate two learning rate schedules: the \emph{conventional decay} schedule (blue), which sets
    $\mathrm{lr}_{\min} = \mathrm{lr}_{\max}/10$, and our
    \emph{fixed} schedule (red), with $\mathrm{lr}_{\min}=10^{-5}$. The visualization reveals
    that the conventional decay method leads to a discernible \textbf{leftward
    bias} in the optimal learning rate range, indicated by the shift of the lowest
    loss region towards lower learning rates in the blue contours compared to
    the red.
    }
    \label{fig:learning_rate}
    \vskip -20pt
\end{wrapfigure}

We investigated two approaches for the final minimum learning rate ($\mathrm{lr}_{\min}$):
the conventional decay schedule ($\mathrm{lr}_{\min} = \mathrm{lr}_{\max}/10$) \cite{Brown2020,Jin2023,Touvron2023a,Touvron2023b,biderman2023pythiasuiteanalyzinglarge,workshop2023bloom176bparameteropenaccessmultilingual,Shen2024}, and proposed fixed schedule ($\mathrm{lr}_{\min} = 10^{-5}$). Training 1B model on 80B tokens, we compared these schedules across various LR and BS.

Fig.~\ref{fig:learning_rate} presents comparative heatmaps of the final training loss. We observe that compared to using a fixed final learning rate, setting it as $\mathrm{lr}_{\max}/10$ shows distinct optimal hyperparameter points and an overall left-skewed distribution of suboptimal learning rate and batch size combinations. We analyze that this is because, for the relatively high peak learning rates, conventional schedules result in disproportionately large minimum learning rates, which adversely affects the final stages of training and prevents the loss from converging to better local optima.
As further illustrated in Fig.~\ref{fig:firt_pic_baselines_compare}, aside from Porian Law, which converges the $\mathrm{lr}_{\min}$ to a sufficiently small value, the optimal learning rates calculated by other traditional learning rate decay schedules all exhibit varying degrees of a left-skew issue.

This aligns with advanced training practices which suggest that the minimum learning rate significantly impacts the loss. This phenomenon is unfavorable for fitting our scaling laws, and in practice, it is generally preferred to keep the $\mathrm{lr}_{\min}$ fixed at a relatively low value. So we adopt the fixed final learning rate strategy in our subsequent experiments.

\subsubsection{Evaluation Metric Consistency}
\label{Evaluation metric Consistency}

\begin{figure*}[!t]
    \centering
    \includegraphics[width=1.0\textwidth,trim=0 0 0 40,clip]{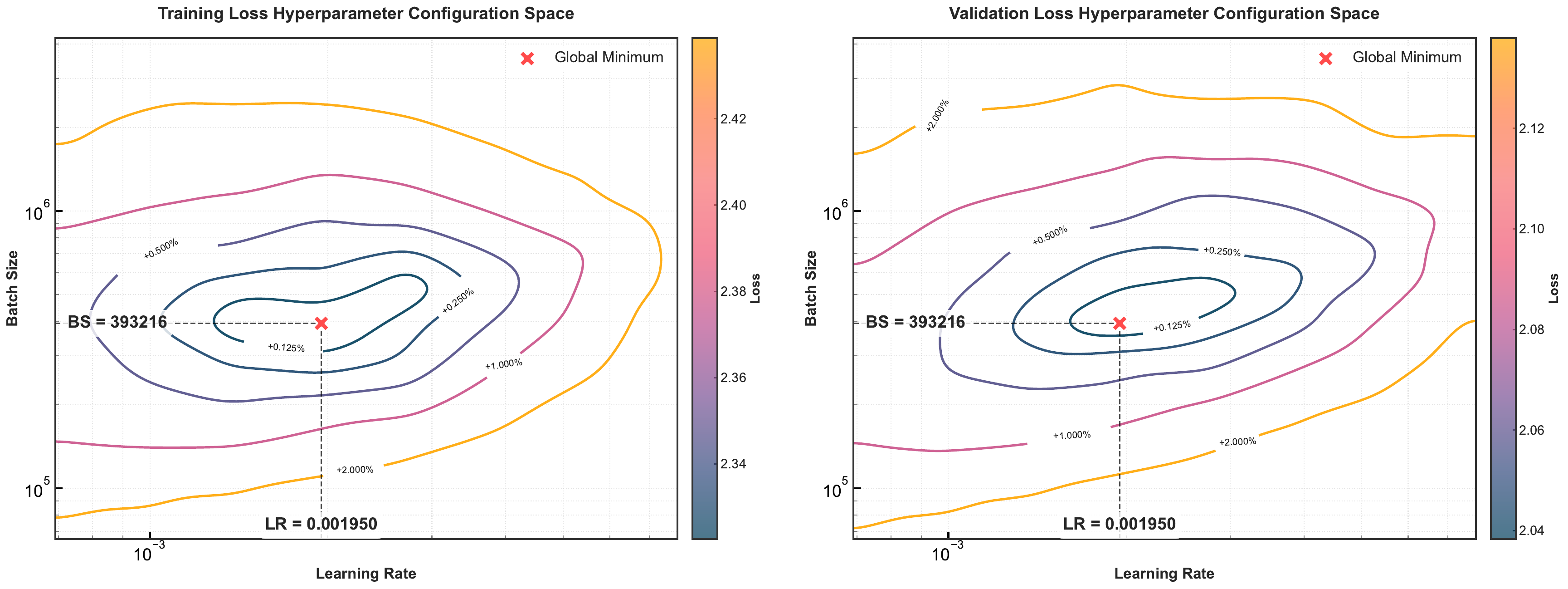}
    \caption{\textbf{Contour plots of training loss (left) and validation loss (right) }. Both plots share the global minimum ({\color{red}\ding{55}}) at batch size 393,216 and learning rate of 0.001950.}
    \vskip -18pt
    \label{fig:loss_bpc_heatmaps}
\end{figure*}

As described in Chinchilla~\cite{hoffmann_training_2022}, smooth training loss
is considered an unbiased estimate of validation loss for simplicity. We operate under this same setting and supplement our investigation with experimental analysis.
As shown in Fig.~\ref{fig:loss_bpc_heatmaps}, for the case where the
smooth training loss converges to the optimal value of $2.279$ ( as indicated
by the solid red-framed line in Fig.~\ref{fig:loss_bpc_heatmaps} right) , the corresponding LR and BS are $1.95\times10^{-3}$ and $393,216$ respectively. This
is the same as the position of the LR and BS corresponding to the validation loss converging to the optimal value of $2.038$ (as indicated by the solid red-framed line
in Fig.~\ref{fig:loss_bpc_heatmaps} left). Moreover, the overall trend of how the smooth training loss deviates from the optimal value with varying learning rates and batch sizes (as shown by the  patterns in Fig.~\ref{fig:loss_bpc_heatmaps} right) closely mirrors the corresponding variations observed in validation loss measurements. This alignment demonstrates that the smooth training loss provides consistent optimization guidance for learning rate and batch size selection, matching the parameter configurations that would be obtained through direct validation loss evaluation.

\subsection{Fitting Hyperparameter Scaling Laws}
\label{fitting_hp_scaling_law}

\subsubsection{Scaling Laws}
\label{LR_BS_wrt_N_D}

\begin{figure*}[!t]
\centering
\begin{subfigure}[t]{0.96\textwidth}
    \includegraphics[width=\linewidth]{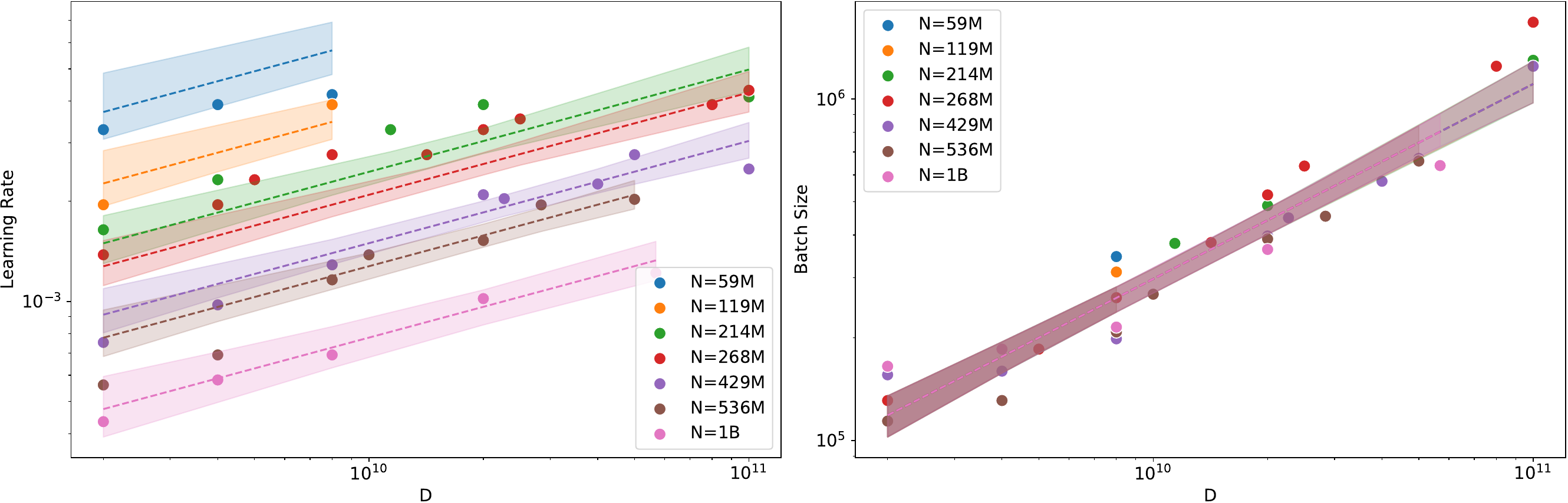}
    \caption{Scaling laws with dataset size for different number of parameters}
    \label{fig:diff_N_LR_BS}
\end{subfigure}
\hfill
\begin{subfigure}[t]{0.96\textwidth}
    \includegraphics[width=\linewidth]{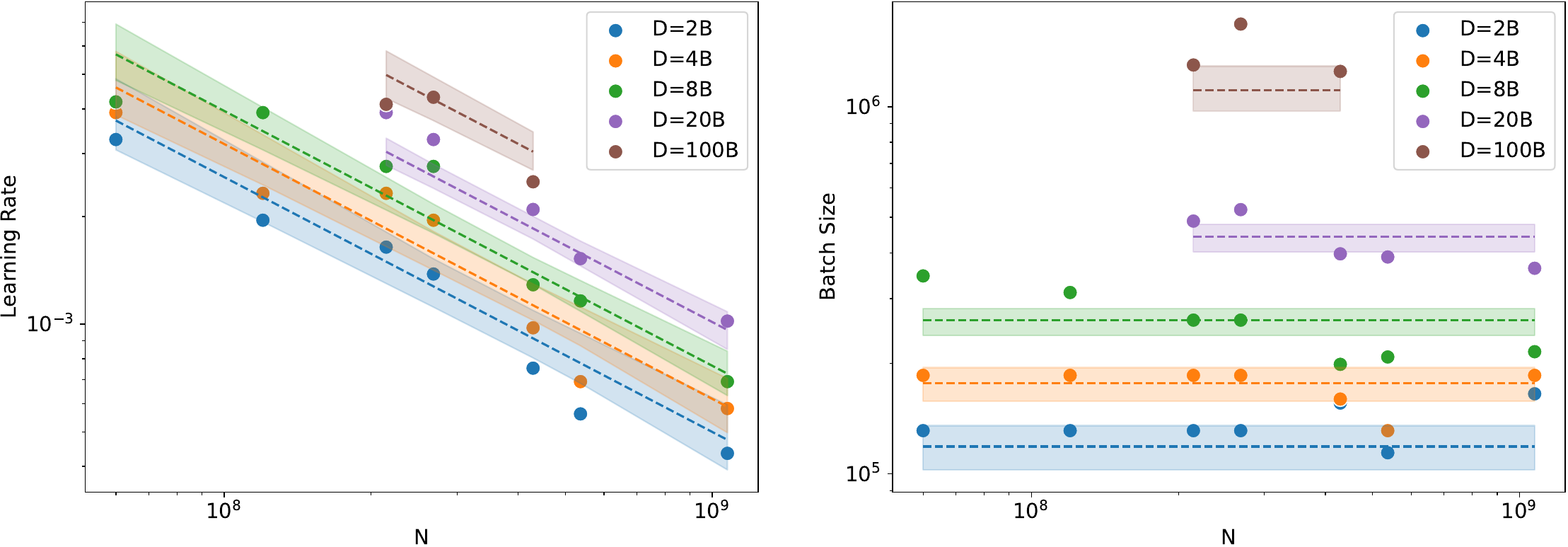}
    \caption{Scaling laws with number of parameters for different dataset size}
    \label{fig:diff_D_LR_BS}
\end{subfigure}
\caption{
\textbf{Empirical vs. predicted HP scaling.}
(a) Scatter points indicate empirical optimal learning rate vs. batch size for model scale $N$;
(b) Analogous results for dataset scale D.
Curves show our hp-scaling law predictions, with shaded regions representing parameter uncertainty bounds from the sampling-based fitting strategy.
Both plots use double logarithmic scaling (1,912 training samples).
}
\vskip -20pt
\label{fig:laws}
\end{figure*}

In accordance with Definition 1, we experimentally derive the LR and BS by keeping
other variables fixed. This section focuses on elucidating the relationships between these empirically determined hyperparameters and $N$ and $D$. For the parameter count $N$, we set up seven experiments spanning 60M, 120M, 210M, 270M, 430M, 540M, and 1B parameters. As demonstrated in Fig.~\ref{fig:diff_N_LR_BS}, our experiments reveal a positive correlation between optimal LR and BS and the data scale $D$ for each value of $N$. Furthermore, we conducted experiments across five different data scales $D$: 2B, 4B, 8B, 20B, and 100B tokens. Notably, we specifically reserved the 1B parameter and 100B token settings as test points to validate our findings, as discussed in Sec.~\ref{sec:validation}.
As visualized in Fig.~\ref{fig:diff_D_LR_BS}, we find that for each data scale
$D$, the optimal LR increases with model size $N$. Notably, our findings
indicate that optimal BS is largely independent of $N$. 

Building upon the above insights, we delve into the scaling behavior of optimal hyperparameters. 
Specifically, we investigate
how the optimal LR scales with $N$ and $D$, and how the optimal BS scales with $D$.
Our empirical observations, particularly when visualized on a log-log scale, reveal a strong linear trend, suggesting a power-law relationship. Based on this, the scaling law for hyperparameters can be
described by the following power-law relationships:
\begin{equation}
\begin{aligned}
    \eta(N, D) & = c N^{\alpha}D^{\beta},  \\
    B(D)       & = d D^{\gamma}
    \label{eq:scaling_law_lr_bs}
\end{aligned}
\end{equation}
\vskip -5pt
where $c, \alpha, \beta, d$, and $\gamma$ are constants,
the values of which will be determined through fitting in Sec.~\ref{fitting_methods}. Notably, the proposed form $B(D)$ assumes that the optimal batch size is independent of $N$. 
This assumption is statistically validated through regression analysis in Appendix~\ref{app:won}.
It is particularly noteworthy that our proposed scaling law demonstrates significant
generality, meaning it is applicable across diverse architectures
$\mathbb{A}$ and data recipes $\mathbb{D}$. This aspect of generality will
be further elaborated upon in Sec.~\ref{Universal_HP_law}.

\subsubsection{Fitting Methodology}
\label{fitting_methods}

\vskip -8pt
\begin{table}[h!]
    \centering
    \footnotesize
    \caption{Fitted power-law coefficients for hyperparameter scaling laws}
    \label{tab:parameters}
    \begin{tabular}{lccccc}
        \toprule \textbf{Parameter}           & $\alpha$ & $\beta$ & $\gamma$ & $c$  & $d$  \\
        \midrule \textbf{Fitted value}        & -0.713   & 0.307   & 0.571    & 1.79 & 0.58 \\
        \bottomrule \label{hp_result}
    \end{tabular}
    \vskip -7pt
\end{table}

Building on Step Law from Sec.~\ref{LR_BS_wrt_N_D}, we
transform the power laws in Eq.~\eqref{eq:scaling_law_lr_bs}
into the linear form:
\vskip -13pt
\begin{align}
    \log \eta & = \log c + \alpha \log N + \beta \log D \label{eq:log_lr} \\
    \log B    & = \log d + \gamma \log D \label{eq:log_bs}
\end{align}
\vskip -3pt

In this way, we can employ Ordinary Least Squares to fit the unknown
parameters $\log c, \alpha, \beta, \log d$ and $\gamma$. Specifically, we
set up 7 groups of experiments with different $N$ and $D$ as shown in Appendix~\ref{app:struct}.
Following~\cite{hoffmann_training_2022}, we fit the optimal LR and BS with
the experimentally predicted LR and BS. We averaged the results of these 1,000 bootstrap samples to obtain the intermediate final parameters. 
This averaged result is what we present in Tab.~\ref{tab:parameters}. 
Furthermore, the variability across these 1,000 bootstrap samples is depicted as the shaded regions in Fig.~\ref{fig:laws}, providing an indication of the uncertainty associated with the fitted results. These shaded regions allow us to visually assess the robustness and confidence of the optimal LR and BS derived from our procedure.

\subsubsection{Comparisons with Existing Approaches}
\label{sec:validation}

With the fitted scaling laws, we directly extrapolate to the test point ($N=1$B, $D=100$B) for comparison. As shown in Fig.~\ref{fig:firt_pic_baselines_compare} and detailed in Appendix~\ref{app:models}, our method consistently finds solutions near the global optimum. In contrast, other approaches show significant deviation.This gap mainly comes from their modeling limitations. 

Previous methods typically fit learning rate  alone, without jointly modeling batch size. For example, DeepSeek Law\cite{DeepSeek-AI2024} assumes a fixed compute budget, which restricts the $(N, D)$ range and degrades fitting accuracy. As discussed in Sec.~\ref{fixed_final_lr}, many approaches fix the final LR as a constant multiple of the initial LR. This often results in overly large final LRs for large initial values, leading to poor convergence. Although Porian Law~\cite{Porian2024} mitigate this with a minimum LR constraint, their method lacks stability. It does not fully account for interactions between hyperparameters and model dimension $D$. This becomes more problematic in MoE (Sec.~\ref{sec:moe}) and data recipe (Sec.~\ref{sec:data}), where small $D/N$ ratios cause their predicted LR and BS to fall outside stable ranges.

We further test on a 6.5B-parameter MoE model with two dataset sizes: $D = 1.0 \times 10^{10}$ and $1.3 \times 10^{11}$ (see configurations 17 and 18 in Appendix~\ref{app:struct}). For each setting, we use Porian Law, DeepSeek Law, and our Step Law to predict LR and BS from $(N, D)$, and train accordingly. We exclude $\mu$P~\cite{Yang2022} as it only predicts LR and does not handle BS.Table~\ref{tab:extended_baselines} shows that our Step Law consistently achieves the lowest validation smooth loss. This confirms that our method generalizes well to large, sparse MoE architectures and remains robust under diverse training regimes.

\vskip -10pt
\begin{table}[h]
\renewcommand{\arraystretch}{0.8}
  \centering
  \small
  \caption{Comparison of predicted HPs and validation smooth loss on a 6.5B MoE model.}
  \label{tab:extended_baselines}
  \begin{tabular}{l c c r}
    \toprule
    Method  & Learning Rate (LR) & Token-wise Batch Size  (BS) & Smooth Loss \\
    \midrule
    \multicolumn{4}{c}{\itshape Dataset size $D = 1.0\times10^{10}$} \\
    \midrule
    \begin{tabular}[c]{@{}l@{}}
      Porian Law 
      \cite{Porian2024}
    \end{tabular}
      & $1.08\times10^{-3}$ 
      & 6029312 
      & 2.4352 \\
    \begin{tabular}[c]{@{}l@{}}
      DeepSeek Law  
      \cite{DeepSeek-AI2024}
    \end{tabular}
      & $1.06\times10^{-3}$ 
      & 851968  
      & 2.2891 \\
    Step Law (Ours)
      & $2.12\times10^{-4}$ 
      & 294912  
      & \textbf{2.2700} \\
    \midrule
    \multicolumn{4}{c}{\itshape Dataset size $D = 1.3\times10^{11}$} \\
    \midrule
    \begin{tabular}[c]{@{}l@{}}
      Porian Law
      \cite{Porian2024}
    \end{tabular}
      & $1.08\times10^{-3}$ 
      & 6029312
      & 1.9772 \\
    \begin{tabular}[c]{@{}l@{}}
      DeepSeek Law
      \cite{DeepSeek-AI2024}
    \end{tabular}
      & $7.70\times10^{-4}$ 
      & 1966080  
      & 1.9511 \\
    Step Law (Ours)
      & $4.74\times10^{-4}$ 
      & 1310720  
      & \textbf{1.9479} \\
    \bottomrule
  \end{tabular}
\end{table}
\vskip -20pt
\subsection{Validation Cross Architecture and Data}
\label{Universal_HP_law}

\subsubsection{Topological Invariance Across Varied Model Shapes}

\begin{figure*}[!t]
    \setlength{\abovecaptionskip}{0pt}
    \setlength{\belowcaptionskip}{10pt}
    \begin{center}
        \centerline{\includegraphics[trim=170 5 170 85, clip, width=0.99\textwidth]{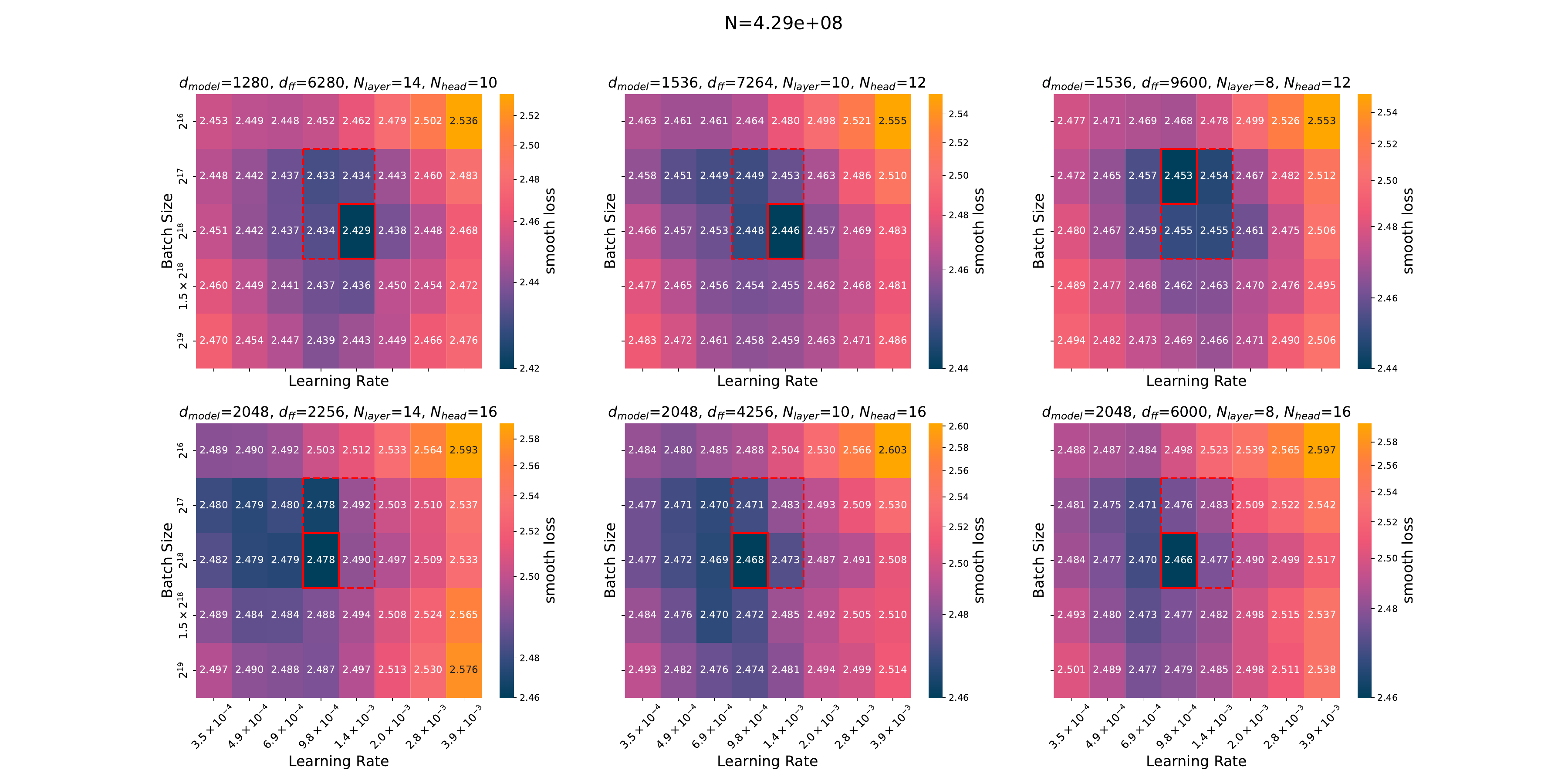}}
        \caption{\textbf{Consistent Hyperparameter Optima Across Diverse Model Topologies.} Smooth loss landscapes across learning rate (LR) and batch size (BS) for six Transformer architectures, each with approximately 430M parameters but differing in shape configurations, varying hidden dimension ($d_{\text{model}}$), feedforward size ($d_{\mathrm{ff}}$), number of layers ($N_{\text{layer}}$), and attention heads ($N_{\text{head}}$). Despite these architectural differences, the optimal LR-BS region (solid box) consistently resides within a narrow band (dashed box), indicating topological invariance in hyperparameter scaling behavior.}
        \label{fig:Topological_Invariance}
        \vskip -35pt
    \end{center}
\end{figure*}

As illustrated in Fig.~\ref{fig:Topological_Invariance}, we conduct a series of controlled experiments to systematically investigate the relationship between HP scaling and model architecture topology. Specifically, we set a model with 430 million parameters and varied its structural configuration by defining six distinct model shape combinations. These model shape variations involved changes in key architectural factors (\eg number of layers, attention heads, feed-forward network dimensions).


For each of the 6 model shapes, we conduct extensive hyperparameter tuning to identify the optimal LR and BS. The results show a clear pattern: across all configurations, the optimal LR and BS consistently fall within a narrow range, despite large variations in model topology. This consistency supports our hypothesis that Step Law is invariant to architectural changes. In particular, variations in depth, attention heads, or feedforward width do not alter the scaling relationships among LR, BS, model size $N$, and dataset size $D$.

\subsubsection{Sparsity-Robustness in MoE Models}
\label{sec:moe}

Step Law has been extensively studied for dense model, but its applicability to sparse architectures remains uncertain. 
MoE~\cite{shazeer2017outrageously,fedus2022switch} activates only a subset of parameters per token, making it structurally different from dense models.
This raises the question of whether Step Law can be generalized to MoE settings. To investigate this, we conducted experiments on MoE models across 16 different sparsity levels and model shapes (refer to Tab.~\ref{tab:moe_struct} in the Appendix~\ref{app:struct}). These settings allow us to examine how the scaling law behaves under different levels of sparsity. We evaluate multiple existing scaling methods under this framework.

\begin{figure*}[h!]
    \centering
    \begin{subfigure}[b]{0.31\textwidth}
        \centering
        \includegraphics[width=\textwidth]{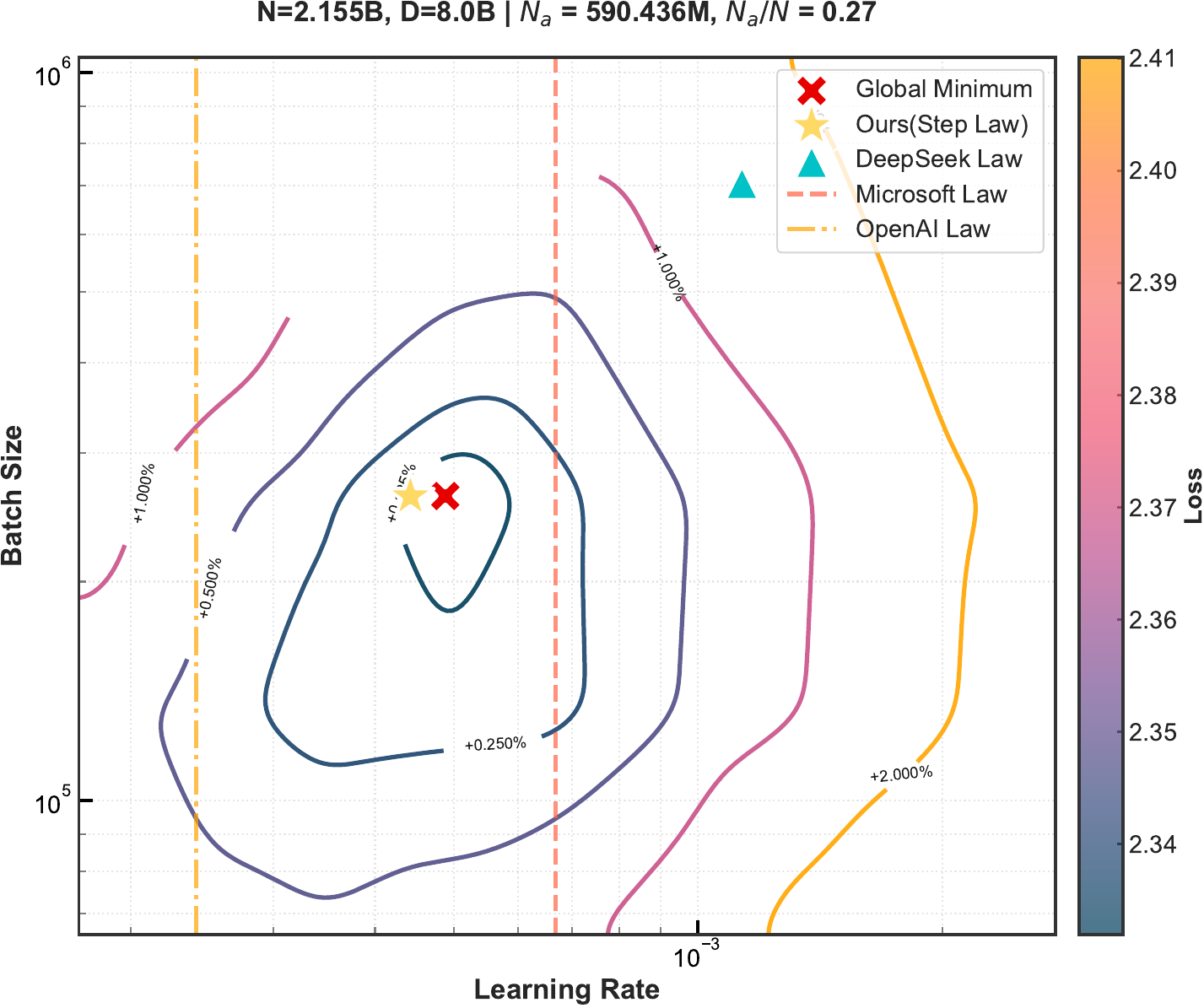}
        \label{fig:moe_subfig1}
    \end{subfigure}
    \hfill
    \begin{subfigure}[b]{0.31\textwidth}
        \centering
        \includegraphics[width=\textwidth]{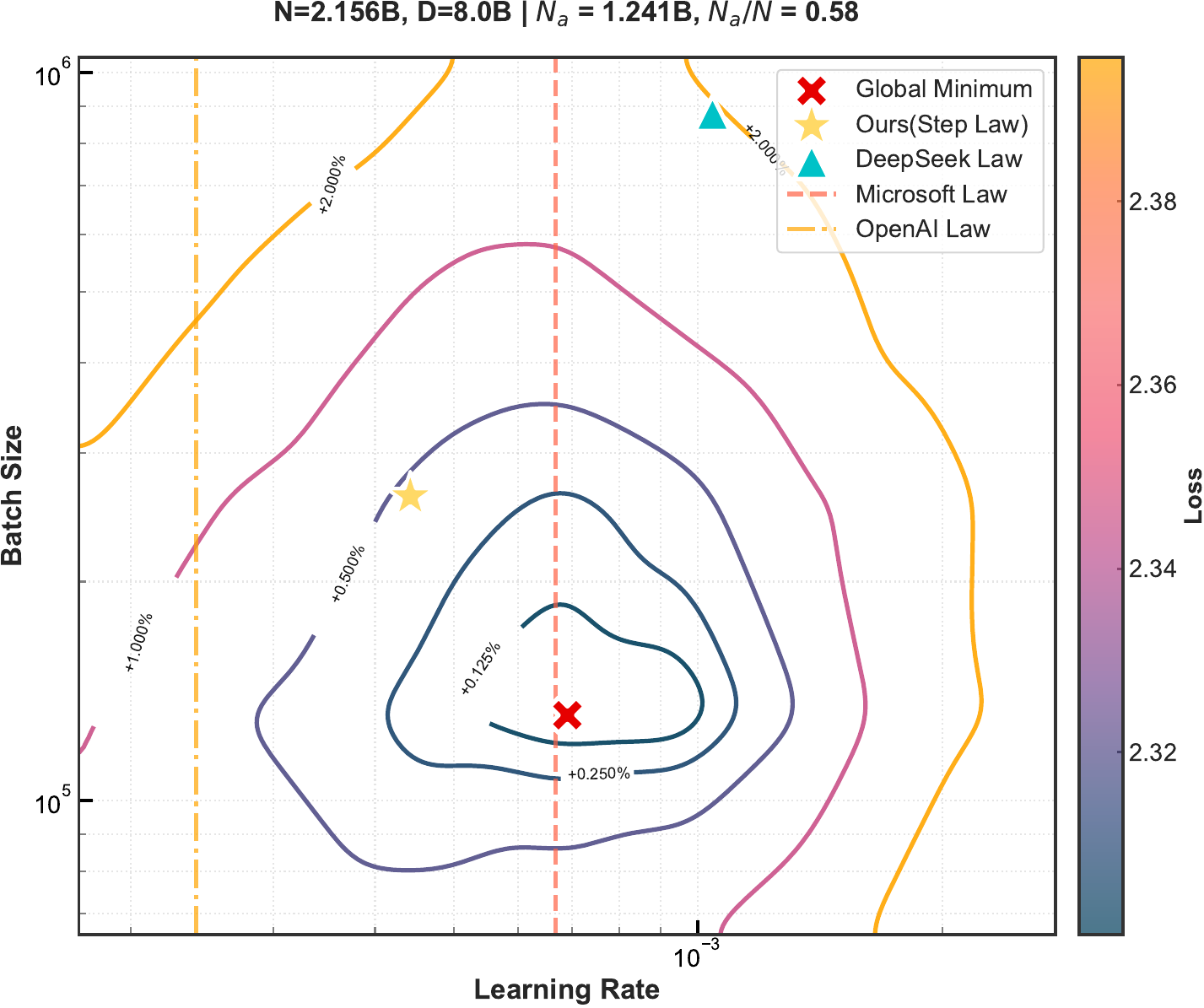}
        \label{fig:moe_subfig2}
    \end{subfigure}
    \hfill
    \begin{subfigure}[b]{0.31\textwidth}
        \centering
        \includegraphics[width=\textwidth]{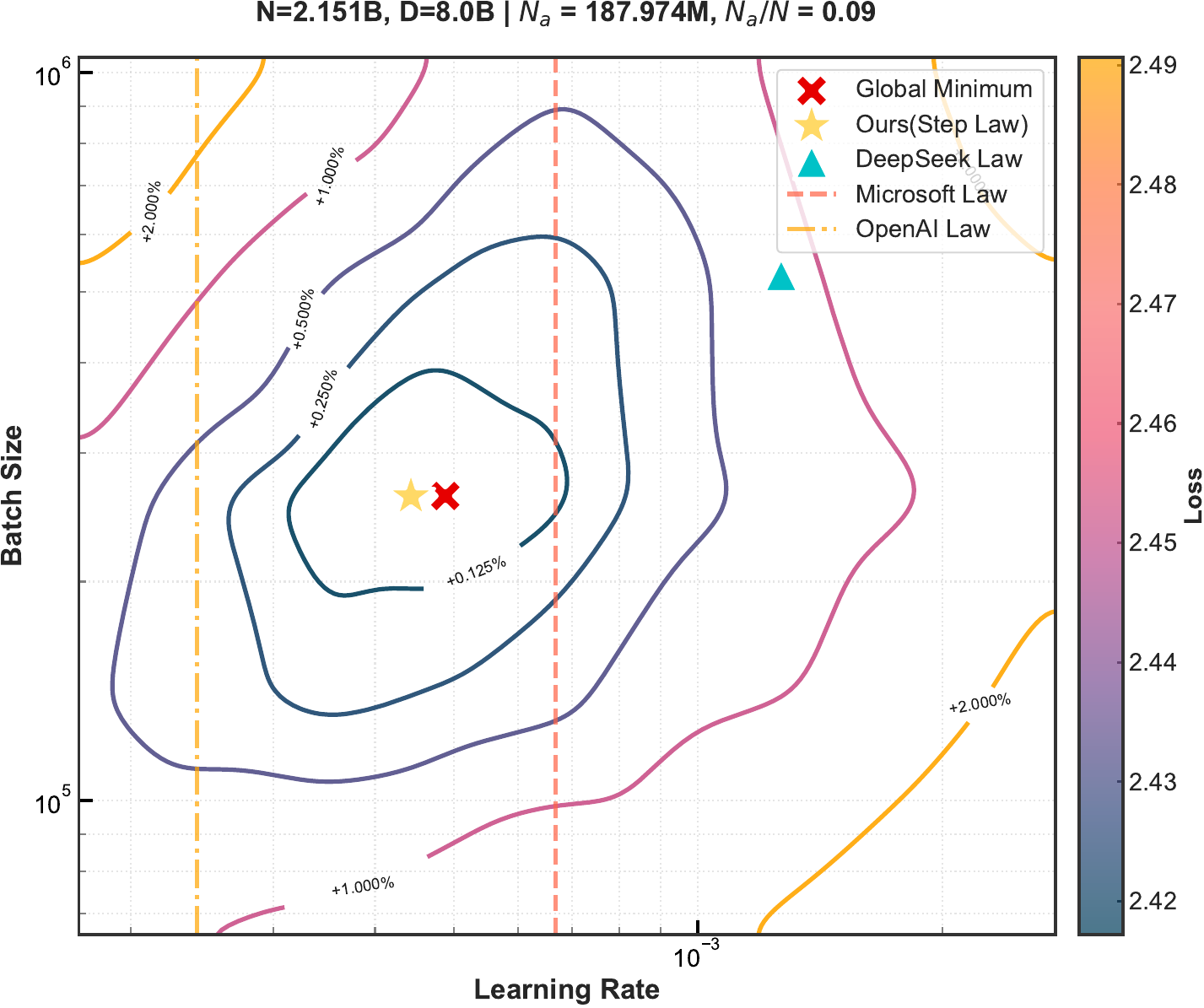}
        \label{fig:moe_subfig3}
    \end{subfigure}

    \vskip -4pt
    \caption{ \textbf{Validation loss landscapes of MoE models under varying sparsity ratios ($ N_a/N $).} Left: Low sparsity ($ N_a/N=0.27 $). Middle: Medium sparsity ($ N_a/N=0.58 $). Right: Medium sparsity at D=8.0B. Our method consistently approximates global minima across sparsity regimes.}
    \vskip -4pt
    \label{fig:moe}
\end{figure*}

As shown in Fig.~\ref{fig:moe}, our approach consistently achieves a relative prediction error within 0.5\% across all sparsity levels, significantly outperforming competing methods. In contrast, the DeepSeek Formula yields a relative error over four times larger, indicating its reduced accuracy in MoE settings. In contrast, our method provides a more comprehensive framework, successfully predicting multiple hyperparameters. 
Additional MoE experiment details and full results across 16 sparsity levels are provided in ~\ref{app:moes}.
These results demonstrate that the Step Law extends beyond dense architectures and remains effective for 
 MoE, regardless of sparsity level. 
This suggests that the underlying principles of scaling laws emerge from broader optimization and capacity constraints rather than being specific to dense parameterization. Our findings reinforce the general applicability of Step Laws and their potential to guide efficient scaling in diverse neural architectures $\mathbb{A}$.

\subsubsection{Data Distribution Generalization}
\label{sec:data}

\begin{figure*}[h!]
    \centering
    \begin{subfigure}[b]{0.31\textwidth}
    \centering
    \includegraphics[width=\textwidth]{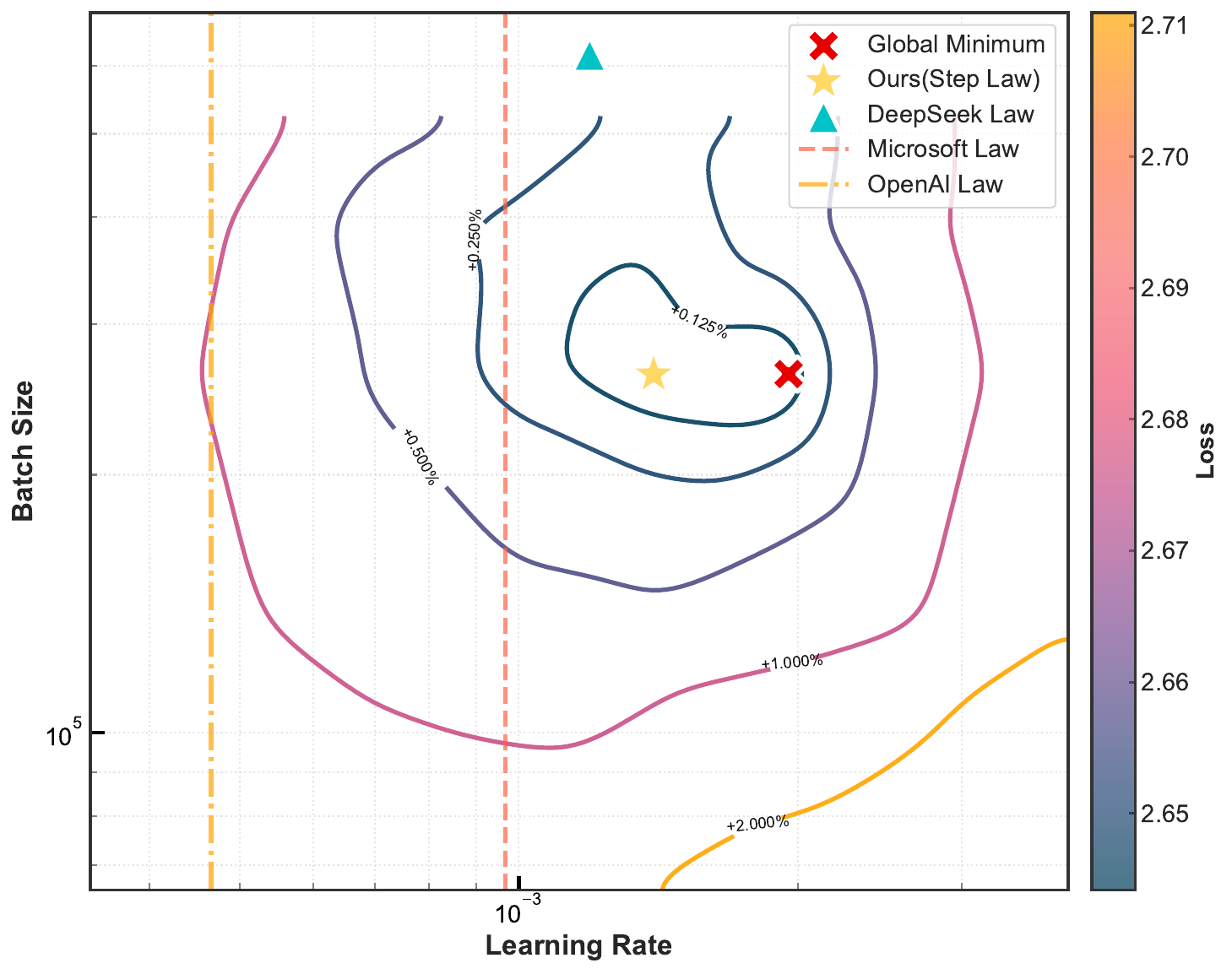}
    \caption{Bilingual Corpus}
    \label{fig:data_subfig1}
    \end{subfigure}
    \hfill
    \begin{subfigure}[b]{0.31\textwidth}
    \centering
    \includegraphics[width=\textwidth]{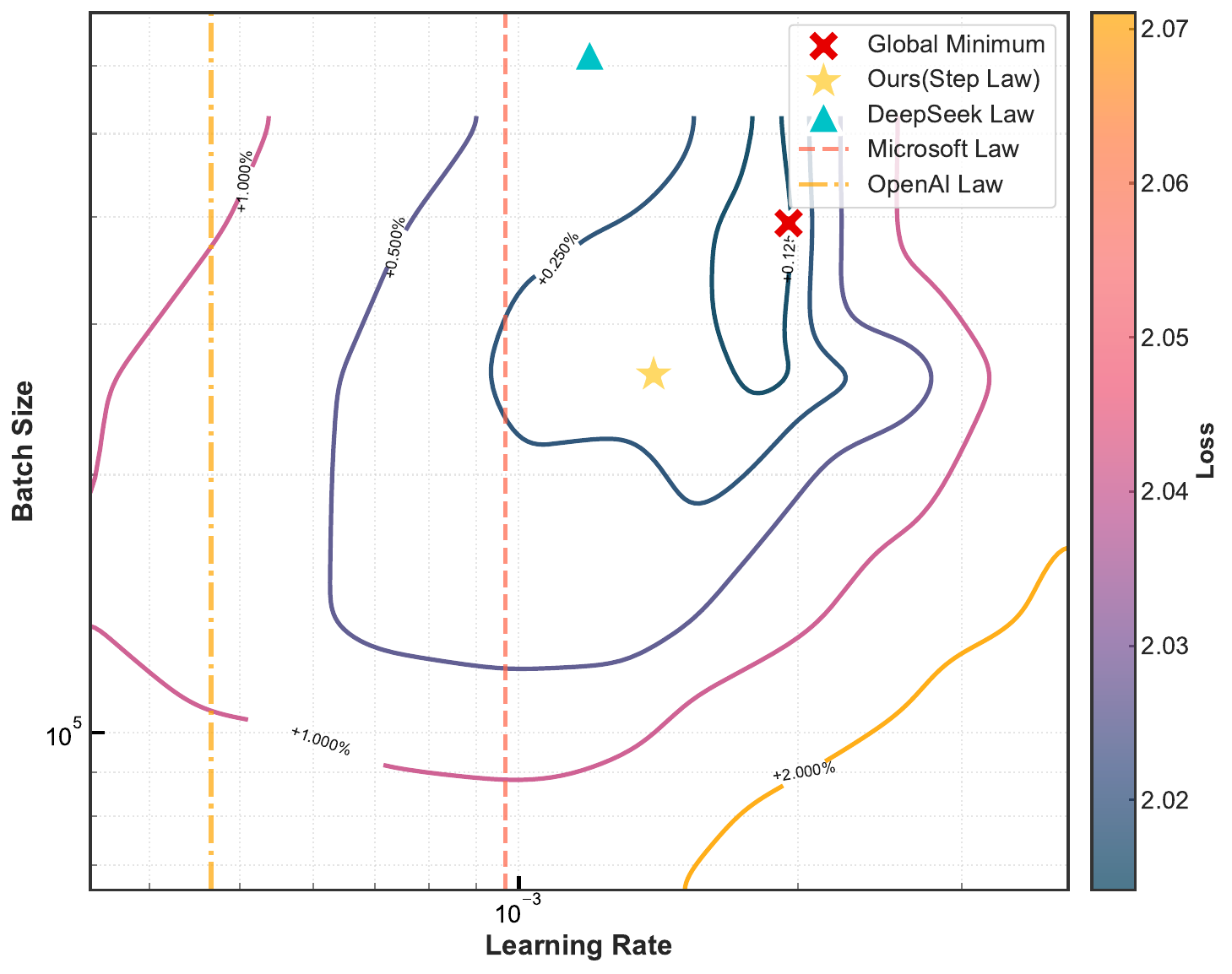}
    \caption{Code Integration}
    \label{fig:data_subfig2}
    \end{subfigure}
    \hfill
    \begin{subfigure}[b]{0.31\textwidth}
    \centering
    \includegraphics[width=\textwidth]{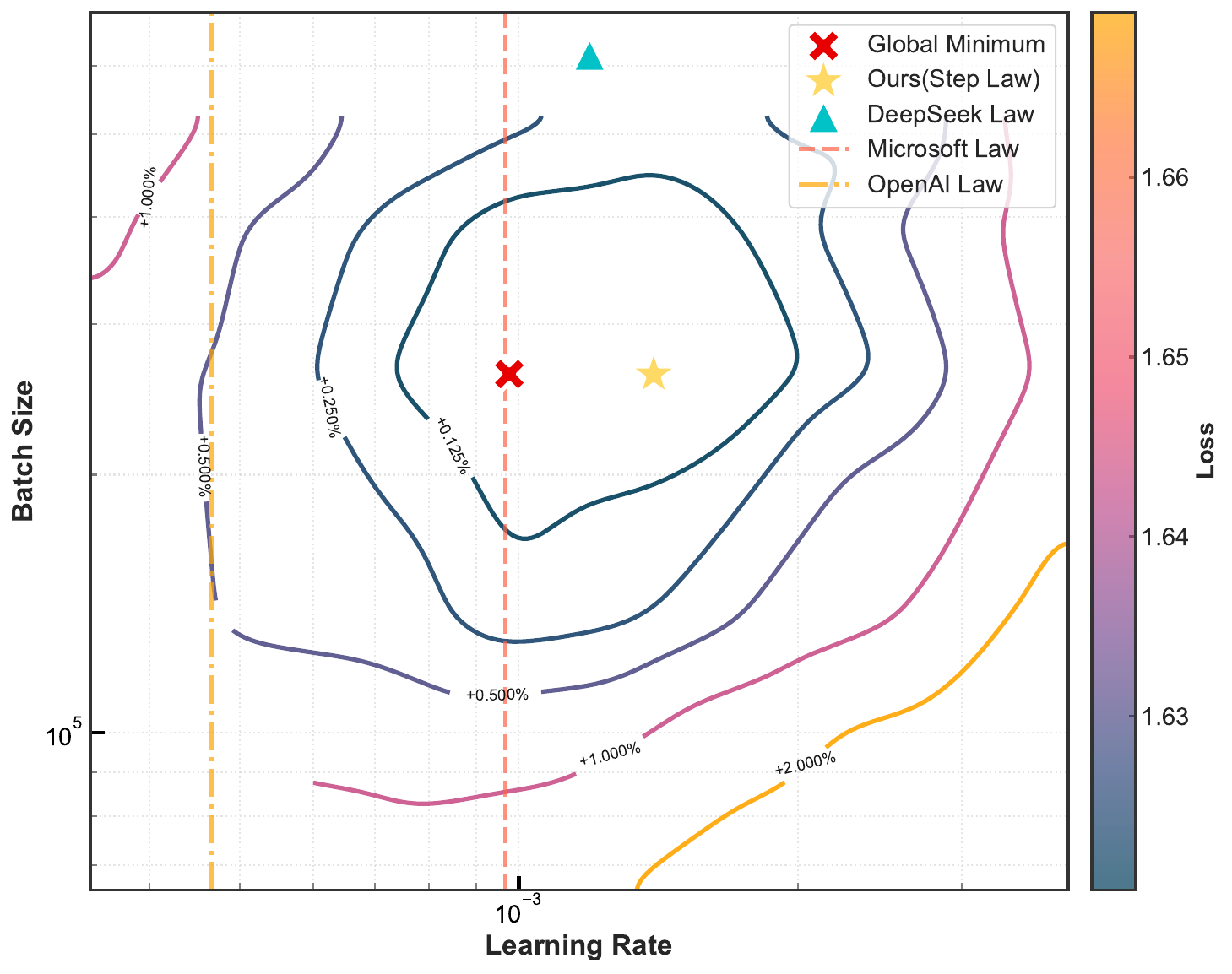}
    \caption{Code-Dominante}
    \label{fig:data_subfig3}
    \end{subfigure}
    \caption{\textbf{Configuration Space Analysis under Different Data Recipes.} Our method demonstrates stable convergence patterns across varying data compositions.}
    \vskip -5pt
\label{fig:data_distribution}
\end{figure*}

To rigorously assess the robustness of our Step Law across varied data recipes $\mathbb{D}$, we design three distinct data recipes, progressively diverging from the original composition, as detailed in Appendix Tab.~\ref{tab:data_recipes}: adding Chinese (Bilingual Corpus), adding code (Code Integration), and mostly code (Code-Dominant).As shown in Fig.~\ref{fig:data_distribution}, our formula achieves high predictive accuracy across all three distributions, with relative errors within 0.25\% of the global minimum. This consistently outperforms alternative methods, which show larger deviations.

These results reveal two key findings:
(a) The Step Law exhibits statistical invariance under both linguistic and structural variations, suggesting strong generalizability beyond standard language distributions;
(b) The predicted optimal hyperparameters remain stable across highly heterogeneous training data, demonstrating the robustness of our approach.
These findings are particularly significant for designing scalable and adaptable training paradigms applicable across diverse deployment scenarios with varying dataset characteristics.

\section{Conclusions}
In this paper, we provide a crucial advancement in efficient hyperparameter
optimization for LLMs. By empirically unveiling and rigorously validating universal
scaling laws for learning rate and batch size—underpinned by the discovery of
loss landscape convexity—we move beyond computationally expensive grid searches
and limited transfer methods. Our robust HP scaling laws, supported by an unprecedentedly
large empirical study and open-sourced resources, empower the community with
a practical and generalizable approach for navigating the hyperparameter
configuration space in LLM pre-training, thereby facilitating more efficient and
scalable LLM development.

\bibliographystyle{plain}
\bibliography{ref_scaling_law.bib}

\begin{thebibliography}{10}

\bibitem{biderman2023pythiasuiteanalyzinglarge}
Stella Biderman, Hailey Schoelkopf, Quentin Anthony, Herbie Bradley, Kyle O'Brien, Eric Hallahan, Mohammad~Aflah Khan, Shivanshu Purohit, USVSN~Sai Prashanth, Edward Raff, Aviya Skowron, Lintang Sutawika, and Oskar van~der Wal.
\newblock Pythia: A suite for analyzing large language models across training and scaling, 2023.

\bibitem{Bjorck2024}
Johan Bjorck, Alon Benhaim, Vishrav Chaudhary, Furu Wei, and Xia Song.
\newblock Scaling optimal lr across token horizons.
\newblock 9 2024.

\bibitem{Blake2024}
Charlie Blake, Constantin Eichenberg, Josef Dean, Lukas Balles, Luke~Y. Prince, Björn Deiseroth, Andres~Felipe Cruz-Salinas, Carlo Luschi, Samuel Weinbach, and Douglas Orr.
\newblock u-$\mu$p: The unit-scaled maximal update parametrization.
\newblock 7 2024.

\bibitem{Bordelon2023}
Blake Bordelon, Lorenzo Noci, Mufan~Bill Li, Boris Hanin, and Cengiz Pehlevan.
\newblock Depthwise hyperparameter transfer in residual networks: Dynamics and scaling limit.
\newblock 9 2023.

\bibitem{Brown2020}
Tom~B. Brown, Benjamin Mann, Nick Ryder, Melanie Subbiah, Jared Kaplan, Prafulla Dhariwal, Arvind Neelakantan, Pranav Shyam, Girish Sastry, Amanda Askell, Sandhini Agarwal, Ariel Herbert-Voss, Gretchen Krueger, Tom Henighan, Rewon Child, Aditya Ramesh, Daniel~M. Ziegler, Jeffrey Wu, Clemens Winter, Christopher Hesse, Mark Chen, Eric Sigler, Mateusz Litwin, Scott Gray, Benjamin Chess, Jack Clark, Christopher Berner, Sam McCandlish, Alec Radford, Ilya Sutskever, and Dario Amodei.
\newblock Language models are few-shot learners.
\newblock 5 2020.

\bibitem{DeepSeek-AI2024}
DeepSeek-AI, Xiao Bi, Deli Chen, Guanting Chen, Shanhuang Chen, Damai Dai, Chengqi Deng, and et~al.
\newblock Deepseek llm: Scaling open-source language models with longtermism.
\newblock 1 2024.

\bibitem{DeepSeek-AI2025}
DeepSeek-AI, Daya Guo, Dejian Yang, Haowei Zhang, Junxiao Song, Ruoyu Zhang, Runxin Xu, Qihao Zhu, Shirong Ma, Peiyi Wang, et~al.
\newblock Deepseek-r1: Incentivizing reasoning capability in llms via reinforcement learning.
\newblock 1 2025.

\bibitem{DeepSeek-AI2024b}
DeepSeek-AI, Aixin Liu, Bei Feng, Bing Xue, Bingxuan Wang, Bochao Wu, Chengda Lu, Chenggang Zhao, Chengqi Deng, Chenyu Zhang, Chong Ruan, et~al.
\newblock Deepseek-v3 technical report.
\newblock {\em arXiv preprint arXiv:2412.19437}, 2024.

\bibitem{Du2021}
Nan Du, Yanping Huang, Andrew~M. Dai, Simon Tong, Dmitry Lepikhin, Yuanzhong Xu, Maxim Krikun, Yanqi Zhou, Adams~Wei Yu, Orhan Firat, Barret Zoph, Liam Fedus, Maarten Bosma, Zongwei Zhou, Tao Wang, Yu~Emma Wang, Kellie Webster, Marie Pellat, Kevin Robinson, Kathleen Meier-Hellstern, Toju Duke, Lucas Dixon, Kun Zhang, Quoc~V Le, Yonghui Wu, Zhifeng Chen, and Claire Cui.
\newblock Glam: Efficient scaling of language models with mixture-of-experts.
\newblock 12 2021.

\bibitem{Everett2024}
Katie Everett, Lechao Xiao, Mitchell Wortsman, Alexander~A. Alemi, Roman Novak, Peter~J. Liu, Izzeddin Gur, Jascha Sohl-Dickstein, Leslie~Pack Kaelbling, Jaehoon Lee, and Jeffrey Pennington.
\newblock Scaling exponents across parameterizations and optimizers.
\newblock 7 2024.

\bibitem{Fedus2021}
William Fedus, Barret Zoph, and Noam Shazeer.
\newblock Switch transformers: Scaling to trillion parameter models with simple and efficient sparsity.
\newblock 1 2021.

\bibitem{fedus2022switch}
William Fedus, Barret Zoph, and Noam Shazeer.
\newblock Switch transformers: Scaling to trillion parameter models with simple and efficient sparsity.
\newblock {\em Journal of Machine Learning Research}, 23(120):1--39, 2022.

\bibitem{Filatov2024}
Oleg Filatov, Jan Ebert, Jiangtao Wang, and Stefan Kesselheim.
\newblock Time transfer: On optimal learning rate and batch size in the infinite data limit.
\newblock 10 2024.

\bibitem{Gage1994ANA}
Philip Gage.
\newblock A new algorithm for data compression.
\newblock {\em The C Users Journal archive}, 12:23--38, 1994.

\bibitem{Grattafiori2024}
Aaron Grattafiori, Abhimanyu Dubey, Abhinav Jauhri, Abhinav Pandey, Abhishek Kadian, Ahmad Al-Dahle, Aiesha Letman, Akhil Mathur, Alan Schelten, Alex Vaughan, and Amy~Yang et~al.
\newblock The llama 3 herd of models.
\newblock 7 2024.

\bibitem{Halfon2024}
Alon Halfon, Shai Gretz, Ofir Arviv, Artem Spector, Orith Toledo-Ronen, Yoav Katz, Liat Ein-Dor, Michal Shmueli-Scheuer, and Noam Slonim.
\newblock Stay tuned: An empirical study of the impact of hyperparameters on llm tuning in real-world applications.
\newblock 7 2024.

\bibitem{hoffmann_training_2022}
Jordan Hoffmann, Sebastian Borgeaud, Arthur Mensch, Elena Buchatskaya, Trevor Cai, Eliza Rutherford, Diego de~Las Casas, Lisa~Anne Hendricks, Johannes Welbl, Aidan Clark, Tom Hennigan, Eric Noland, Katie Millican, George van~den Driessche, Bogdan Damoc, Aurelia Guy, Simon Osindero, Karen Simonyan, Erich Elsen, Jack~W. Rae, Oriol Vinyals, and Laurent Sifre.
\newblock Training {Compute}-{Optimal} {Large} {Language} {Models}, March 2022.
\newblock arXiv:2203.15556 [cs].

\bibitem{Hu2024}
Shengding Hu, Yuge Tu, Xu~Han, Chaoqun He, Ganqu Cui, Xiang Long, Zhi Zheng, Yewei Fang, Yuxiang Huang, Weilin Zhao, Xinrong Zhang, Zheng~Leng Thai, Kaihuo Zhang, Chongyi Wang, Yuan Yao, Chenyang Zhao, Jie Zhou, Jie Cai, Zhongwu Zhai, Ning Ding, Chao Jia, Guoyang Zeng, Dahai Li, Zhiyuan Liu, and Maosong Sun.
\newblock Minicpm: Unveiling the potential of small language models with scalable training strategies.
\newblock 4 2024.

\bibitem{Jin2023}
Hongpeng Jin, Wenqi Wei, Xuyu Wang, Wenbin Zhang, and Yanzhao Wu.
\newblock Rethinking learning rate tuning in the era of large language models.
\newblock 9 2023.

\bibitem{kaplan_scaling_2020}
Jared Kaplan, Sam McCandlish, Tom Henighan, Tom~B. Brown, Benjamin Chess, Rewon Child, Scott Gray, Alec Radford, Jeffrey Wu, and Dario Amodei.
\newblock Scaling {Laws} for {Neural} {Language} {Models}, January 2020.
\newblock arXiv:2001.08361 [cs, stat].

\bibitem{Lingle2024}
Lucas Lingle.
\newblock A large-scale exploration of $\mu$-transfer.
\newblock 4 2024.

\bibitem{Loshchilov2017}
Ilya Loshchilov and Frank Hutter.
\newblock Decoupled weight decay regularization.
\newblock 11 2017.

\bibitem{Ludziejewski2025}
Jan Ludziejewski, Maciej Pióro, Jakub Krajewski, Maciej Stefaniak, Michał Krutul, Jan Małaśnicki, Marek Cygan, Piotr Sankowski, Kamil Adamczewski, Piotr Miłoś, and Sebastian Jaszczur.
\newblock Joint moe scaling laws: Mixture of experts can be memory efficient.
\newblock 2 2025.

\bibitem{McCandlish2018}
Sam McCandlish, Jared Kaplan, Dario Amodei, and OpenAI~Dota Team.
\newblock An empirical model of large-batch training.
\newblock 12 2018.

\bibitem{Perko2023}
Stefan Perko.
\newblock Unlocking optimal batch size schedules using continuous-time control and perturbation theory.
\newblock 12 2023.

\bibitem{Porian2024}
Tomer Porian, Mitchell Wortsman, Jenia Jitsev, Ludwig Schmidt, and Yair Carmon.
\newblock Resolving discrepancies in compute-optimal scaling of language models.
\newblock 6 2024.

\bibitem{Press2021}
Ofir Press, Noah~A. Smith, and Mike Lewis.
\newblock Train short, test long: Attention with linear biases enables input length extrapolation.
\newblock 8 2021.

\bibitem{workshop2023bloom176bparameteropenaccessmultilingual}
Teven~Le Scao, Angela Fan, Christopher Akiki, Ellie Pavlick, Suzana Ilić, Daniel Hesslow, Roman Castagné, Alexandra~Sasha Luccioni, François Yvon, Matthias Gallé, Thomas Wolf, et~al.
\newblock Bloom: A 176b-parameter open-access multilingual language model.
\newblock 11 2022.

\bibitem{Shazeer2020}
Noam Shazeer.
\newblock Glu variants improve transformer.
\newblock 2 2020.

\bibitem{shazeer2017outrageously}
Noam Shazeer, Azalia Mirhoseini, Krzysztof Maziarz, Andy Davis, Quoc Le, Geoffrey Hinton, and Jeff Dean.
\newblock Outrageously large neural networks: The sparsely-gated mixture-of-experts layer.
\newblock {\em arXiv preprint arXiv:1701.06538}, 2017.

\bibitem{Shen2024}
Yikang Shen, Matthew Stallone, Mayank Mishra, Gaoyuan Zhang, Shawn Tan, Aditya Prasad, Adriana~Meza Soria, David~D. Cox, and Rameswar Panda.
\newblock Power scheduler: A batch size and token number agnostic learning rate scheduler.
\newblock 8 2024.

\bibitem{Huawei-Noah-Ark-Lab2024}
Xian Shuai, Yiding Wang, Yimeng Wu, Xin Jiang, and Xiaozhe Ren.
\newblock Scaling law for language models training considering batch size.
\newblock {\em arXiv preprint arXiv:2412.01505}, 2024.

\bibitem{JMLR:v15:srivastava14a}
Nitish Srivastava, Geoffrey Hinton, Alex Krizhevsky, Ilya Sutskever, and Ruslan Salakhutdinov.
\newblock Dropout: A simple way to prevent neural networks from overfitting.
\newblock {\em Journal of Machine Learning Research}, 15(56):1929--1958, 2014.

\bibitem{Touvron2023a}
Hugo Touvron, Thibaut Lavril, Gautier Izacard, Xavier Martinet, Marie-Anne Lachaux, Timothée Lacroix, Baptiste Rozière, Naman Goyal, Eric Hambro, Faisal Azhar, Aurelien Rodriguez, Armand Joulin, Edouard Grave, and Guillaume Lample.
\newblock Llama: Open and efficient foundation language models.
\newblock 2 2023.

\bibitem{Touvron2023b}
Hugo Touvron, Louis Martin, Kevin Stone, Peter Albert, Amjad Almahairi, Yasmine Babaei, Nikolay Bashlykov, Soumya Batra, Prajjwal Bhargava, and Shruti~Bhosale et~al.
\newblock Llama 2: Open foundation and fine-tuned chat models.
\newblock 7 2023.

\bibitem{Wang2024}
Siqi Wang, Zhengyu Chen, Bei Li, Keqing He, Min Zhang, and Jingang Wang.
\newblock Scaling laws across model architectures: A comparative analysis of dense and moe models in large language models.
\newblock pages 5583--5595, 10 2024.

\bibitem{Wen2024}
Kaiyue Wen, Zhiyuan Li, Jason Wang, David Hall, Percy Liang, and Tengyu Ma.
\newblock Understanding warmup-stable-decay learning rates: A river valley loss landscape perspective.
\newblock 10 2024.

\bibitem{Yang2024}
An~Yang, Baosong Yang, Binyuan Hui, Bo~Zheng, Bowen Yu, Chang Zhou, Chengpeng Li, Chengyuan Li, Dayiheng Liu, Fei Huang, Guanting Dong, Qwen Team, and Alibaba~Group et~al.
\newblock Qwen2 technical report.
\newblock 2024.

\bibitem{Yang2020}
Greg Yang and Edward~J. Hu.
\newblock Feature learning in infinite-width neural networks.
\newblock 11 2020.

\bibitem{Yang2022}
Greg Yang, Edward~J. Hu, Igor Babuschkin, Szymon Sidor, Xiaodong Liu, David Farhi, Nick Ryder, Jakub Pachocki, Weizhu Chen, and Jianfeng Gao.
\newblock Tensor programs v: Tuning large neural networks via zero-shot hyperparameter transfer.
\newblock 3 2022.

\bibitem{Yang2023}
Greg Yang, Dingli Yu, Chen Zhu, and Soufiane Hayou.
\newblock Tensor programs vi: Feature learning in infinite-depth neural networks.
\newblock 10 2023.

\bibitem{Zhang2019}
Biao Zhang and Rico Sennrich.
\newblock Root mean square layer normalization.
\newblock 10 2019.

\bibitem{zhang2024}
Hanlin Zhang, Depen Morwani, Nikhil Vyas, Jingfeng Wu, Difan Zou, Udaya Ghai, Dean Foster, and Sham Kakade.
\newblock How does critical batch size scale in pre-training?
\newblock 10 2024.

\end{thebibliography}

\newpage
    
\appendix
\section{Appendix}
\label{sec:appendix}

\subsection{Notation}
\label{sec:notation}
We use the following notation:
\begin{itemize}
\item $\mathcal{L}$: Cross-entropy loss.
\item $D$: Dataset size in tokens.
\item $N$: Number of non-embedding parameters in the model.
\item $\hat{N}$: Total number of parameters in the model, which excludes embedding layer but includes the model\text{'}s head layer.
\item $C$: Compute budget in FLOPs.
\item $N_{\text{layer}}$: Number of layers in the Transformer model.
\item $d_{\text{ff}}$: Dimension of the feed-forward network hidden layer in the Transformer.
\item $d_{\text{model}}$: Hidden dimension of the Transformer model.
\item $N_{\text{head}}$: Number of attention heads in the Transformer model.
\item $\text{LR}$: Learning rate.
\item $\text{BS}$: Batch size (in tokens).
\item $\eta(N, D)$: Optimal peak learning rate for a given parameter count $N$ and dataset size $D$.
\item $B(N, D)$: Optimal batch size (in tokens) for given parameter $N$ and dataset size $D$.
\item $\mathbb{A}$: Model architecture space defined by $N_{\text{layer}}$, $d_{\text{ff}}$, $d_{\text{model}}$, and $N_{\text{head}}$.
\item $\mathbb{D}$: Training data distribution governed by the data-generating probability distribution.
\end{itemize}

\subsection{Model Scale Dominates Optimal Hyperparameter Selection Over
Computational Complexity}

\begin{table*}[b!]
    \centering
    \resizebox{\textwidth}{!}{
    \begin{tabular}{ccccccccc}
        \hline
        $d_{{model}}$ & $d_{{ff}}$  & $N_{{head}}$ & $N_{{layer}}$ & $\eta(N, D)$       & $B(N, D)$  & $D$        & $N$        & $M$        \\
        \hline
        \multicolumn{9}{c}{Constant $N$ Experiments} \\
        \hline
        1280 & 12264 & 10 & 8  & $1.95 \times 10^{-3}$ & $262{,}144$ & $8.00 \times 10^9$  & $4.29 \times 10^8$ & $2.83 \times 10^9$ \\
        1280 & 6280  & 10 & 14 & $1.38 \times 10^{-3}$ & $262{,}144$ & $8.00 \times 10^9$  & $4.29 \times 10^8$ & $3.02 \times 10^9$ \\
        1536 & 9600  & 12 & 8  & $9.77 \times 10^{-4}$ & $131{,}072$ & $8.00 \times 10^9$  & $4.29 \times 10^8$ & $2.88 \times 10^9$ \\
        1536 & 7264  & 12 & 10 & $1.38 \times 10^{-3}$ & $262{,}144$ & $8.00 \times 10^9$  & $4.29 \times 10^8$ & $2.95 \times 10^9$ \\
        1536 & 4608  & 12 & 14 & $9.77 \times 10^{-4}$ & $131{,}072$ & $8.00 \times 10^9$  & $4.29 \times 10^8$ & $3.10 \times 10^9$ \\
        2048 & 6000  & 16 & 8  & $9.77 \times 10^{-4}$ & $262{,}144$ & $8.00 \times 10^9$  & $4.29 \times 10^8$ & $2.98 \times 10^9$ \\
        2048 & 4256  & 16 & 10 & $9.77 \times 10^{-4}$ & $262{,}144$ & $8.00 \times 10^9$  & $4.29 \times 10^8$ & $3.08 \times 10^9$ \\
        2048 & 2256  & 16 & 14 & $9.77 \times 10^{-4}$ & $262{,}144$ & $8.00 \times 10^9$  & $4.29 \times 10^8$ & $3.28 \times 10^9$ \\
        \hline
        \multicolumn{9}{c}{Constant $M$ Experiments} \\
        \hline
        1280 & 12608 & 10 & 8  & $1.38 \times 10^{-3}$ & $262{,}144$ & $8.00 \times 10^9$  & $4.40 \times 10^8$ & $2.89 \times 10^9$ \\
        1280 & 5888  & 10 & 14 & $1.38 \times 10^{-3}$ & $262{,}144$ & $8.00 \times 10^9$  & $4.08 \times 10^8$ & $2.89 \times 10^9$ \\
        1536 & 9656  & 12 & 8  & $1.38 \times 10^{-3}$ & $262{,}144$ & $8.00 \times 10^9$  & $4.31 \times 10^8$ & $2.89 \times 10^9$ \\
        1536 & 7040  & 12 & 10 & $1.38 \times 10^{-3}$ & $262{,}144$ & $8.00 \times 10^9$  & $4.19 \times 10^8$ & $2.89 \times 10^9$ \\
        1536 & 4056  & 12 & 14 & $9.77 \times 10^{-4}$ & $262{,}144$ & $8.00 \times 10^9$  & $3.94 \times 10^8$ & $2.89 \times 10^9$ \\
        2048 & 5704  & 16 & 8  & $9.77 \times 10^{-4}$ & $262{,}144$ & $8.00 \times 10^9$  & $4.15 \times 10^8$ & $2.89 \times 10^9$ \\
        2048 & 3744  & 16 & 10 & $6.91 \times 10^{-4}$ & $131{,}072$ & $8.00 \times 10^9$  & $3.98 \times 10^8$ & $2.89 \times 10^9$ \\
        2048 & 1504  & 16 & 14 & $6.91 \times 10^{-4}$ & $131{,}072$ & $8.00 \times 10^9$  & $3.64 \times 10^8$ & $2.89 \times 10^9$ \\
        \hline
    \end{tabular}
    }
    \caption{\textbf{Model configurations and results for constant $N$ and constant $M$ experiments.}
    The first group (top) maintains constant parameter count $N$ $\approx$ $4.29 \times 10^8$,
    while the second group (bottom) maintains constant computational complexity
    $M$ $\approx$ $2.89 \times 10^9$. $M$:  non-embedding FLOPs/token.}
    \label{tab:model_configs}
\end{table*}

To investigate how model architecture variations affect optimal hyperparameter
settings, we conducted two sets of control experiments. In the first set, we
maintained a constant parameter count ($N$), while in the second set, we kept the computational complexity ($M$) constant. Both sets used identical training configurations with 8B training tokens, varying only in their architectural
proportions.

Tab.~\ref{tab:model_configs} presents the detailed configurations and
results for both experimental groups. For each model, we systematically
varied the hidden dimension ($d_{{model}}$), feed-forward dimension ($d_{{ff}}$), number of attention heads ($N_{{head}}$), and number of layers ($N_{{layer}}$) while maintaining either constant $N$ or $M$. The embedding dimension ($D$) was fixed at $8.00 \times 10^9$ across all experiments.

\begin{figure*}[!t]
    \setlength{\abovecaptionskip}{0pt}
    \setlength{\belowcaptionskip}{10pt}
    \begin{center}
        \centerline{\includegraphics[trim=170 5 170 85, clip,width=1.0\textwidth]{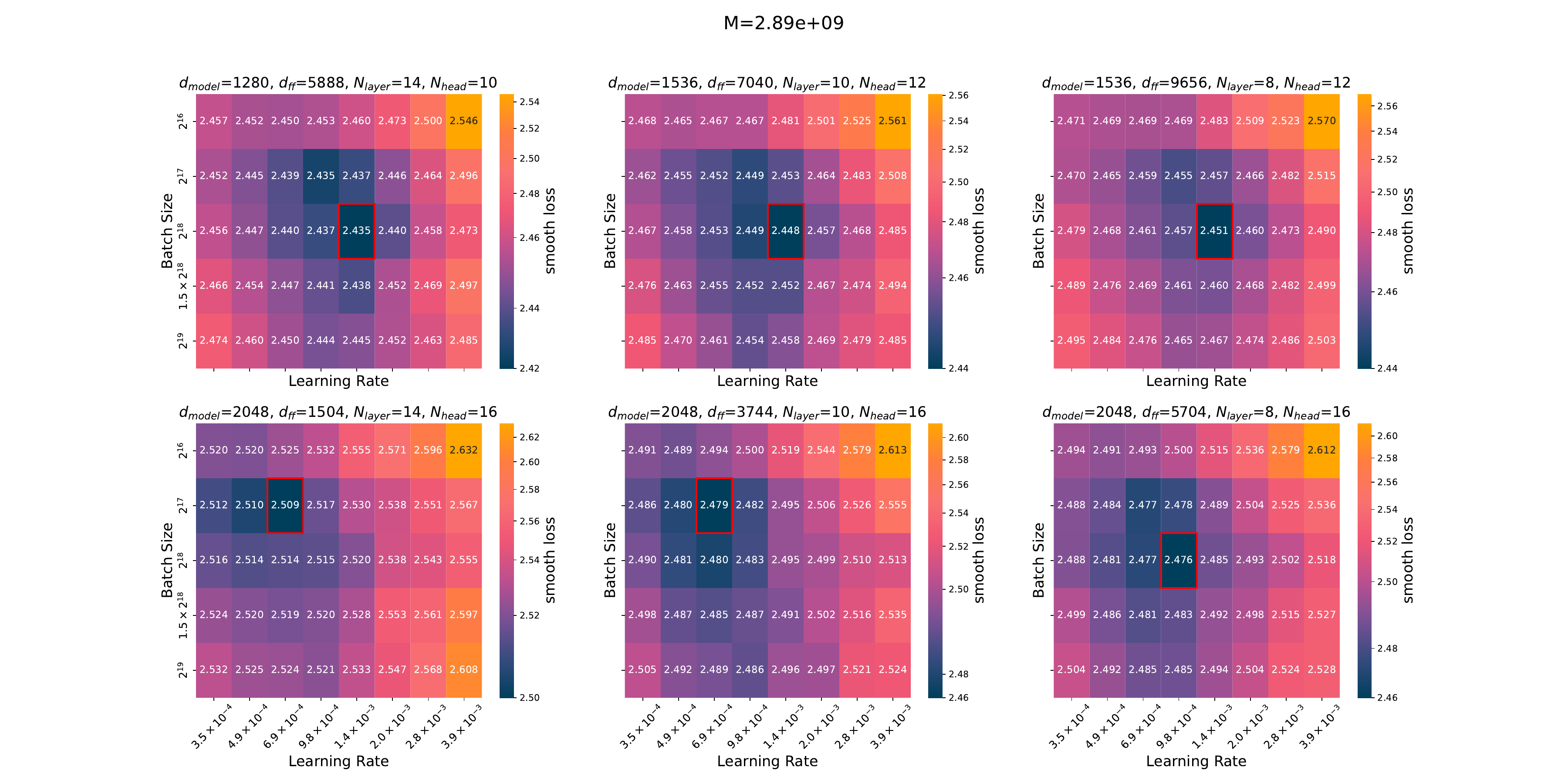}}  
        \caption{ Loss landscapes visualized as heatmaps across learning rate (x-axis)
    and batch size (y-axis) configurations. Darker colors indicate lower loss
    values. Shows results for models with constant
    computational complexity $M$, exhibiting slightly more variance in optimal
    hyperparameter regions.}
        \label{fig:M_shape}
    \end{center}
\end{figure*}

To visualize the impact of hyperparameters across different architectural configurations,
we generated heatmaps of the loss landscape with respect to LR and BS in Fig. \ref{fig:Topological_Invariance} and
\ref{fig:M_shape}. The heatmaps reveal consistent patterns in the optimal
hyperparameter regions across different architectural configurations within
each experimental group.

The experimental results reveal several key findings: (\rmnum{1}) Models with
constant $N$ demonstrate remarkably consistent optimal hyperparameter regions,
with minimal variation in minimum loss values (ranging from 2.4294 to 2.4776)
despite significant architectural differences. (\rmnum{2}) The constant M
experiments show slightly more variation in optimal hyperparameter regions
and minimum loss values (ranging from 2.4346 to 2.5089), suggesting that parameter
count $N$ may be a more robust indicator for hyperparameter selection than computational
complexity $M$. (\rmnum{3}) Across both experimental groups, the optimal
learning rates typically fall within a narrow range ($6.91 \times 10^{-4}$ to $1.95 \times {-3}$), and
batch sizes cluster around either 131,072 or 262,144, regardless of the specific
architectural configuration.

These findings strongly suggest that the fundamental scale metrics,
particularly the parameter count $N$, are more influential in determining optimal
hyperparameter settings than specific architectural choices. This
observation motivates our discussion of hyperparameter scaling laws in
relation to $N$ in Sec.~\ref{fitting_hp_scaling_law}.

\subsection{Model Structural Parameters}
\label{app:struct}
Table~\ref{tab:struct} and Table~\ref{tab:moe_struct} summarizes the precise architectural settings for all models evaluated in our study. In the \textbf{dense model} group (Models~1-18), we cover models ranging from $2.15\times10^{8}$ to $1.07\times10^{9}$ total parameters by varying the hidden dimension ($d_{\mathrm{model}}$), feed‑forward width ($d_{\mathrm{ff}}$), number of attention heads ($N_{\mathrm{head}}$) and number of layers ($N_{\mathrm{layer}}$). Each entry also lists the corresponding dataset size $D$ used during training. In the \textbf{MoE model} group (Models~1-16), we hold the overall parameter count fixed at approximately $2.15\times10^{9}$ while sweeping dataset size $D$ from $2\times10^{9}$ up to $2\times10^{10}$. We further vary the number of experts ($N_{\mathrm{expert}}$), per‑expert hidden size ($d_{\mathrm{moe}}$), top‑$k$ routing, and the resulting active parameter count ($N_a$). This systematic variation allows direct comparison of dense versus sparse Mixture‑of‑Experts architectures under matched compute budgets and data scales.

\begin{table*}[h!]
    \centering
    \begin{tabular}{ccccccc}
        \toprule Model & {$N$} & {$D$} & $d_{{model}}$ & $d_{{ff}}$ & $N_{{head}}$ & $N_{{layer}}$ \\
        \midrule 
        1 
& $2.15 \times 10^8$ & $1.14 \times 10^{10}$ & 960 & 9368 & 15 & 7 \\
        2 
& $4.29 \times 10^8$ & $5.00 \times 10^{10}$ & 1280 & 9472 & 10 & 10 \\
        3 
& $2.68 \times 10^8$ & $8.00 \times 10^{10}$ & 1024 & 9552 & 16 & 8 \\
        4 
& $4.29 \times 10^8$ & $8.00 \times 10^9$ & 1280 & 9472 & 10 & 10 \\
        5 
& $1.07 \times 10^9$ & $2.00 \times 10^{10}$ & 2048 & 8192 & 16 & 16 \\
        6 
& $5.37 \times 10^8$ & $1.00 \times 10^{10}$ & 1280 & 9048 & 10 & 13 \\
        7 
& $2.15 \times 10^8$ & $4.00 \times 10^9$ & 960 & 9368 & 15 & 7 \\
        8 
& $2.68 \times 10^8$ & $5.00 \times 10^9$ & 1024 & 9552 & 16 & 8 \\
        9 
& $2.68 \times 10^8$ & $1.42 \times 10^{10}$ & 1024 & 9552 & 16 & 8 \\
        10 
& $1.07 \times 10^9$ & $5.69 \times 10^{10}$ & 2048 & 8192 & 16 & 16 \\
        11 
& $2.15 \times 10^8$ & $1.00 \times 10^{11}$ & 960 & 9368 & 15 & 7 \\
        12 
& $4.29 \times 10^8$ & $2.27 \times 10^{10}$ & 1280 & 9472 & 10 & 10 \\
        13 
& $5.37 \times 10^8$ & $2.84 \times 10^{10}$ & 1280 & 9048 & 10 & 13 \\
        14 
& $2.15 \times 10^8$ & $2.00 \times 10^{10}$ & 960 & 9368 & 15 & 7 \\
        15 
& $4.29 \times 10^8$ & $4.00 \times 10^{10}$ & 1280 & 9472 & 10 & 10 \\
        16 
& $2.68 \times 10^8$ & $2.50 \times 10^{10}$ & 1024 & 9552 & 16 & 8 \\
        17 & $5.37 \times 10^8$ & $5.00 \times 10^{10}$ & 1280 & 9048 & 10 & 13 \\
        18& $1.07 \times 10^9$ & $1.00 \times 10^{11}$ & 2048 & 8192 & 16 & 16 \\
        \bottomrule
    \end{tabular}
    \caption{\textbf{Dense Model Configuration.}} 
    \label{tab:struct}
\end{table*}


\begin{table*}
[h!]
\centering
\resizebox{\textwidth}{!}{
\begin{tabular}{cccccccccc}
\toprule
Model & {$N$} & {$D$} & {$d_{{model}}$} & {$N_{{head}}$} & {$N_{{layer}}$} & {$N_{{expert}}$} & {$d_{{moe}}$} & {Top-$k$} & {$N_a$} \\
\midrule
    1  
& $2.151 \times 10^9$  & $2.00 \times 10^9$  & 1408 & 11 & 16 & 89 & 352  & 1 & $1.88 \times 10^8$ \\
    2  
& $2.151 \times 10^9$  & $2.00 \times 10^9$  & 1408 & 11 & 16 & 88 & 352  & 2 & $2.33 \times 10^8$ \\
    3  
& $2.155 \times 10^9$  & $2.00 \times 10^9$  & 1408 & 11 & 16 & 8  & 3528 & 1 & $5.90 \times 10^8$ \\
    4  
& $2.156 \times 10^9$  & $2.00 \times 10^9$  & 1408 & 11 & 16 & 8  & 2888 & 3 & $1.24 \times 10^9$ \\
    5  
& $2.151 \times 10^9$  & $4.00 \times 10^9$  & 1408 & 11 & 16 & 89 & 352  & 1 & $1.88 \times 10^8$ \\
    6  
& $2.151 \times 10^9$  & $4.00 \times 10^9$  & 1408 & 11 & 16 & 88 & 352  & 2 & $2.33 \times 10^8$ \\
    7  
& $2.155 \times 10^9$  & $4.00 \times 10^9$  & 1408 & 11 & 16 & 8  & 3528 & 1 & $5.90 \times 10^8$ \\
    8  
& $2.156 \times 10^9$  & $4.00 \times 10^9$  & 1408 & 11 & 16 & 8  & 2888 & 3 & $1.24 \times 10^9$ \\
    9  
& $2.151 \times 10^9$  & $8.00 \times 10^9$  & 1408 & 11 & 16 & 89 & 352  & 1 & $1.88 \times 10^8$ \\
    10 
& $2.151 \times 10^9$  & $8.00 \times 10^9$  & 1408 & 11 & 16 & 88 & 352  & 2 & $2.33 \times 10^8$ \\
    11 
& $2.155 \times 10^9$  & $8.00 \times 10^9$  & 1408 & 11 & 16 & 8  & 3528 & 1 & $5.90 \times 10^8$ \\
    12 
& $2.156 \times 10^9$  & $8.00 \times 10^9$  & 1408 & 11 & 16 & 8  & 2888 & 3 & $1.24 \times 10^9$ \\
    13 
& $2.151 \times 10^9$  & $2.00 \times 10^{10}$ & 1408 & 11 & 16 & 89 & 352  & 1 & $1.88 \times 10^8$ \\
    14 
& $2.151 \times 10^9$  & $2.00 \times 10^{10}$ & 1408 & 11 & 16 & 88 & 352  & 2 & $2.33 \times 10^8$ \\
    15 & $2.155 \times 10^9$  & $2.00 \times 10^{10}$ & 1408 & 11 & 16 & 8  & 3528 & 1 & $5.90 \times 10^8$ \\
    16& $2.156 \times 10^9$  & $2.00 \times 10^{10}$ & 1408 & 11 & 16 & 8  & 2888 & 3 & $1.24 \times 10^9$ \\
    17& $6.510 \times 10^9$  & $1.00 \times 10^{10}$ & 2048 & 16 & 24 & 82  & 512 & 2 & $7.26 \times 10^8$ \\
    18& $6.510 \times 10^9$  & $1.30 \times 10^{11}$ & 2048 & 16 & 24 & 82  & 512 & 2 & $7.26 \times 10^8$ \\
\bottomrule
\end{tabular}
}
        \caption{\textbf{MoE Model Configuration.} {$N_{{expert}}$} denotes the number of experts. {$d_{{moe}}$} denotes the hidden size of experts. 
 {Top-$k$} denotes the number in the routing algorithm. {$N_a$} denotes the activate parameters.} 
\label{tab:moe_struct}
\end{table*}

\FloatBarrier

\subsection{Composition of Training Datasets} %
\label{tab:data_recipes}
Tab.~\ref{tab:datasets} details the dataset weight percentages for four training recipes: Baseline, Code+Math, More Code+Math, and En-CN (English-Chinese bilingual).

\begin{table*}[h!]
\small
\centering
\begin{tabular}{lccccc}
\toprule
\textbf{Dataset} & \textbf{Baseline} & \textbf{Code+Math} & \textbf{More Code+Math} & \textbf{En-CN} \\
\midrule
Web-data-en & \centering 79.53 & 44.75 & 20.00 & 44.99 \\
Web-data-cn & -- & -- & -- & 34.52 \\
Code-the-stack & 4.62 & 32.36 & 57.05 & 4.63 \\
Web-data-math & -- & 7.07 & 7.07 & -- \\
Book-non-novel-en & 4.35 & 4.34 & 4.34 & 4.35 \\
Paper & 3.38 & 3.37 & 3.37 & 3.38 \\
Wikipedia-mtlg & 3.24 & 3.24 & 3.24 & 3.25 \\
Stackexchange & 2.21 & 2.21 & 2.21 & 2.22 \\
Wikipedia-en & 1.69 & 1.69 & 1.69 & 1.69 \\
Book-novel-en & 0.83 & 0.83 & 0.83 & 0.83 \\
Wikipedia-cn & 0.13 & 0.13 & 0.13 & 0.13 \\
\bottomrule
\end{tabular}
\caption{\textbf{Comparison of dataset weights (\%) across different training recipes.} Each recipe represents a different focus: baseline, enhanced code and mathematics capability, and English-Chinese bilingual ability. }
\label{tab:datasets}
\end{table*}
\textbf{Bilingual Corpus:} We augmented the original English-only dataset with Chinese data, creating a bilingual distribution to test the law's validity in multilingual settings.\\
\textbf{Code Integration:} We reduced English content and incorporated 32.36\% of the code-the-stack dataset, examining the law's adaptability to code-heavy distributions.\\
\textbf{Code-Dominant:} We further decreased English content and increased code-the-stack to 57.05\%, representing an extreme shift towards code-based data.

\FloatBarrier

\subsection{Statistical Validation of Batch Size Scaling Relationships}
\label{app:won}
\paragraph{Empirical validation.}

This part provides a statistical analysis to examine the claim in the main text that batch size $B$ is independent of model parameter count $N$ but dependent on training dataset size $D$. We conducted a multivariate regression analysis on all experimental configurations. All variables are log-transformed to linearize power-law relationships (logB, logN, logD), consistent with the main text.  Three regression formulations were compared:

\begin{align*}
\text{N-only Formulation:}     \quad & \log B = \beta_0 + \beta_1 \log N \\
\text{D-only Formulation:}     \quad & \log B = \beta_0 + \beta_2 \log D \\
\text{Full Formulation:}       \quad & \log B = \beta_0 + \beta_1 \log N + \beta_2 \log D
\end{align*}

We fit each of these models using ordinary least squares (OLS) regression and perform hierarchical F-tests to assess the contribution of $\log N$ and $\log D$ to predicting $\log B$. The results are summarized in Table ~\ref{tab:nd} and Table~\ref{tab:coeffs} .

\begin{table}[h]
\centering
\caption{Regression formulation comparisons}
\label{tab:nd}
\begin{tabular}{@{}cccc@{}}
\toprule
\textbf{Comparison} & \textbf{Adj. $R^2$} & \textbf{$\Delta R^2$ VS Full} & \textbf{F-statistic} \\
\midrule
N-only  & -0.032 & -85.4\% & 0.08 \\
D-only & 0.821 & -0.2\% & 138.2  \\
Full & 0.823 & 0.0\% & 70.59  \\
\bottomrule
\end{tabular}
\end{table}

\begin{table}[h]
\centering
\caption{Coefficient analysis for Full formulation}
\label{tab:coeffs}
\begin{tabular}{@{}cccccc@{}}
\toprule
\textbf{Predictor} & \textbf{Coeff.} & \textbf{Std. Error} & \textbf{t-value} & \textbf{p-value} & \textbf{95\% CI} \\
\midrule
$\log N$ & -0.087 & 0.075 & -1.158 & 0.257 & [-0.241, 0.067] \\
$\log D$ & 0.580 & 0.049 & 11.863 & $<$0.001 & [0.480, 0.680] \\
\bottomrule
\end{tabular}
\end{table}

 As shown in Appendix Table~\ref{tab:nd} and Table~\ref{tab:coeffs}, the D-only formulation achieves nearly identical explanatory power as the full model (\(R^2 = 0.821\) vs. \(0.823\)), while the N-only model performs poorly (\(R^2 < 0\)). Moreover, in the full model, the coefficient of \(\log D\) is highly significant (\(p < 0.001\)), whereas that of \(\log N\) is not (\(p = 0.257\)). These results confirm that the optimal batch size scales with $D$, but not with $N$.

\subsection{Loss Landscape Convexity}
\label{app:convex}
Experiments are carried out on 18 unique model architectures, with their precise parameter settings summarized in Appendix~\ref{tab:struct}. For each model, the Smooth Loss metric is derived through a grid search over the hyperparameters of learning rate and batch size. Visualization of the smooth loss landscapes substantiates the claims in the main text, demonstrating a consistent convex relationship between the hyperparameters (learning rate and batch size) and the training loss across all models. These findings confirm the robustness and stability of the optimization process as asserted in the primary discussion.

\foreach \fA/\iA/\fB/\iB in {
  ND_214663680_4000000000_3D_convex/0/ND_214663680_11400000000_3D_convex/1,
  ND_214663680_20000000000_3D_convex/2/ND_214663680_100000000000_3D_convex/3,
  ND_268304384_5000000000_3D_convex/4/ND_268304384_14200000000_3D_convex/5,
  ND_268304384_25000000000_3D_convex/6/ND_268304384_80000000000_3D_convex/7,
  ND_429260800_8000000000_3D_convex/8/ND_429260800_22700000000_3D_convex/9,
  ND_429260800_40000000000_3D_convex/10/ND_429260800_50000000000_3D_convex/11,
  ND_536872960_10000000000_3D_convex/12/ND_536872960_28400000000_3D_convex/13,
  ND_536872960_50000000000_3D_convex/14/ND_1073741824_20000000000_3D_convex/15,
  ND_1073741824_56900000000_3D_convex/16/ND_1073741824_100000000000_3D_convex/17
}{
  \pgfmathparse{int(\iA+1)} \let\iAa=\pgfmathresult
  \pgfmathparse{int(\iB+1)} \let\iBb=\pgfmathresult

  \begin{figure}[h!]
    \centering
    \begin{subfigure}[t]{0.48\textwidth}
      \centering
      \includegraphics[trim=0 0 0 100,clip,width=\linewidth]{convex_plots_3d/\fA.pdf}
      \caption{Model \iAa}
      \label{fig:convex_\iA}
    \end{subfigure}
    \hfill
    \begin{subfigure}[t]{0.48\textwidth}
      \centering
      \includegraphics[trim=0 0 0 100,clip,width=\linewidth]{convex_plots_3d/\fB.pdf}
      \caption{Model \iBb}
      \label{fig:convex_\iB}
    \end{subfigure}
    \caption{Illustration of Hyperparameter Configuration Spaces for Models \iAa\ and \iBb.}
    \label{fig:convex_\iA_\iB}
  \end{figure}
}
\FloatBarrier


\subsection{Dense Models}
\label{app:models}
This section presents the full set of hyperparameter configuration space visualizations for the 18 dense Transformer models described in Table~\ref{tab:struct}. Each plot illustrates the validation loss surface as a function of learning rate (LR) and batch size (BS), using a log–log scale for both axes. These visualizations reveal consistent trends across model scales, including the emergence of convex, bowl-shaped minima and smooth shifts in optimal hyperparameter regions. They serve as empirical evidence for the predictive structure captured by the scaling law, and demonstrate its robustness across a wide range of dense model sizes. 

\foreach \iA/\iB in {
    0/1, 2/3, 4/5, 6/7,
    8/9, 10/11, 12/13, 14/15, 16/17%
}{
  \pgfmathparse{int(\iA+1)} \let\iAa=\pgfmathresult
  \pgfmathparse{int(\iB+1)} \let\iBb=\pgfmathresult
  
  \begin{figure}[h!]
    \centering
    \begin{subfigure}[t]{0.48\textwidth}
      \centering
      \includegraphics[width=\linewidth]{compare_pics/\iA.pdf}
      \caption{Model \iAa}
      \label{fig:compare_\iA}
    \end{subfigure}
    \hfill
    \begin{subfigure}[t]{0.48\textwidth}
      \centering
      \includegraphics[width=\linewidth]{compare_pics/\iB.pdf}
      \caption{Model \iBb}
      \label{fig:compare_\iB}
    \end{subfigure}
    \caption{Illustration of Hyperparameter Configuration Spaces for Models \iAa\ and \iBb.}
    \label{fig:compare_\iA_\iB}
  \end{figure}
}
\FloatBarrier


\subsection{MoE Models}
\label{app:moes}
To assess the generality of HP scaling laws beyond dense Transformers, we conduct a comprehensive study on MoE models, which activate only a subset of experts per token. We evaluate 16 distinct configurations (see Table~\ref{tab:moe_struct}), varying both total parameter count and sparsity. For each model, we sweep learning rate (LR) and batch size (BS) over the same logarithmic grid used in our dense experiments.

\foreach \iA/\iB in {
  0/1, 2/3, 4/5, 6/7,
  8/9, 10/11, 12/13, 14/15%
}{
  \pgfmathparse{int(\iA+1)} \let\iAa=\pgfmathresult
  \pgfmathparse{int(\iB+1)} \let\iBb=\pgfmathresult

  \begin{figure}[h!]
    \centering
    \begin{subfigure}[t]{0.48\textwidth}
      \centering
      \includegraphics[width=\linewidth]{moe_pics/moe_\iA.pdf}
      \caption{MoE Model \iAa}
      \label{fig:moe_\iA}
    \end{subfigure}
    \hfill
    \begin{subfigure}[t]{0.48\textwidth}
      \centering
      \includegraphics[width=\linewidth]{moe_pics/moe_\iB.pdf}
      \caption{MoE Model \iBb}
      \label{fig:moe_\iB}
    \end{subfigure}
    \caption{Illustration of Hyperparameter Configuration Spaces for MoE Models \iAa\ and \iBb.}
    \label{fig:moe_\iA_\iB}
  \end{figure}
}
\FloatBarrier

\subsection{Limitations}
\label{sec:limitation} While our empirical study provides valuable universal
HP scaling laws and demonstrates their practical efficacy, it is essential
to acknowledge the limitations inherent in an empirical approach. Our findings
are primarily data-driven. Future work should focus on developing a more
theoretical understanding of the observed power-law relationships, potentially
deriving them from first principles to enhance their predictive power and generalizability
beyond the empirically validated domain.

\end{document}